\documentclass[10pt,twocolumn,letterpaper]{article}

\usepackage{cvpr}
\usepackage{times}
\usepackage{epsfig}
\usepackage{graphicx}
\usepackage{amsmath}
\usepackage{amssymb}
\usepackage{graphicx}
\usepackage{csvsimple}
\usepackage{multirow}
\usepackage[table,xcdraw]{xcolor}
\usepackage{caption}
\usepackage{pdfpages}
\usepackage{subcaption}
\usepackage{multirow}
\usepackage{amsmath,amssymb} 
\usepackage{color}
\usepackage{algorithmic}
\usepackage{algorithm}
\usepackage{framed}
\usepackage{adjustbox}
\usepackage{nicefrac}
\usepackage{url}
\usepackage{authblk}

\DeclareMathOperator*{\argmin}{{\arg\!\min}}

\usepackage{wrapfig}

\newcommand{\z}{\mathbf{z}}
\newcommand{\trace}{\mathrm{trace}}
\usepackage[normalem]{ulem}

\usepackage[pagebackref=true,breaklinks=true,letterpaper=true,colorlinks,bookmarks=false]{hyperref}

 \cvprfinalcopy 

\def\cvprPaperID{1981} 

\newcommand{\comment}[1]{{ }}
\ifcvprfinal\fi
\begin{document}


\title{Solving Uncalibrated Photometric Stereo Using Fewer Images by Jointly Optimizing Low-rank Matrix Completion and Integrability}


\author[1]{Soumyadip Sengupta} 
\author[1]{ Hao Zhou}
\author[2]{Walter  Forkel}
\author[3]{Ronen  Basri}
\author[4] { Tom Goldstein}
\author[1]{David W. Jacobs} 
\affil[1]{Center for Automation Research, University of Maryland, College Park, MD, USA.} 
\affil[2]{TU Dresden, Germany.}
\affil[3]{Department of Computer Science and Applied Mathematics, Weizmann Institute of Science, Rehovot, Israel}
\affil[4]{Department of Computer Science, University of Maryland, College Park, MD, USA.}

\maketitle

\begin{abstract}
   We introduce a new, integrated approach to uncalibrated photometric stereo. We perform 3D reconstruction of Lambertian objects using multiple images produced by unknown, directional light sources.  We show how to formulate a single optimization that includes rank and integrability constraints, allowing also for missing data.  We then solve this optimization using the Alternate Direction Method of Multipliers (ADMM). We conduct extensive experimental evaluation on real and synthetic data sets. Our integrated approach is particularly valuable when performing photometric stereo using as few as 4-6 images, since the integrability constraint is capable of improving estimation of the linear subspace of possible solutions. We show good improvements over prior work in these cases.

\end{abstract}

\section{Introduction}
Uncalibrated photometric stereo (UPS) is the problem of recovering the 3D shape of an object and associated lighting conditions, given images taken with varying, unknown illumination. In this work we replace the existing pipeline for solving UPS with an integrated approach. This paper, like much prior work ~\cite{hayakawa,gbr,yuille,wu,alldrin,papa12,papa13}, focuses on Lambertian objects illuminated by a single distant point light source in each image. Existing methods, pioneered by~\cite{hayakawa}, formulate UPS as the problem of finding a low-rank factorization of the measurements. Specifically, given $m$ images each with $p$ pixels, let $M$ denote the $m \times p$ matrix containing the pixel intensities. These methods optimize
\begin{equation}  \label{eq:rank3}
        \underset{\hat{M}}{\min} \|\hat{M}-M \|^{2}_{F} \quad
        \mathrm{s.t. \quad rank}(\hat{M})=3.
\end{equation}
This problem can be solved by SVD, from which we produce a family of solutions, each consisting of a set of light sources, albedos, and surface normals.
These solutions are related by a $3 \times 3$ ambiguity matrix. The surface normals provided by SVD are in general inconsistent with the partial derivatives of the surface (i.e. they are not \textit{integrable}). Consequently, existing methods apply an additional sequence of steps aimed at reducing the ambiguity and fitting a surface to the recovered normals.

In this paper we propose instead to optimize:
\begin{eqnarray}
\label{eq:ourcost}
&\underset{\hat{M}}{\min} \|\hat{M}-M \|^{2}_{F}  \quad \\ \nonumber
\mathrm{s.t.}  &\hat{M} \ \text{{is rank 3 and produced by an integrable surface}}.
\end{eqnarray}Eq.~\eqref{eq:rank3} optimizes over rank 3 matrices, which can represent sets of images produced by any set of surface normals.  In contrast, in~\eqref{eq:ourcost} we optimize over only those rank 3 matrices that correspond to integrable surfaces.

\begin{figure}[t]
        \centering
        \includegraphics[width=.23\textwidth]{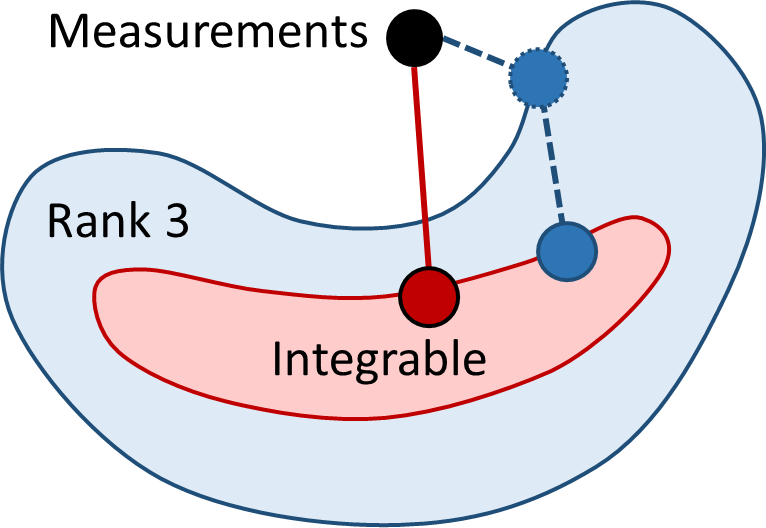}
        \caption{\small A cartoon of our approach.  Blue represents the set of rank 3 matrices, while red represents the subset of those that correspond to integrable surfaces. Our optimization seeks to find the integrable matrix (red dot) that is closest to the measurements (black dot).  If instead we first find the nearest rank 3 matrix and then select an integrable matrix (the blue dots) we may produce a suboptimal solution. }
        \label{fig:illustration}
\end{figure}

Intuitively, a single optimization over all constraints will have a better global optimum than a sequence of optimizations in which constraints are used one at a time to increasingly narrow the solution (see illustration in Figure \ref{fig:illustration}). A similar intuition has motivated the use of bundle adjustment \cite{Hartley2004} as the dominant approach to large scale structure-from-motion. Specifically in UPS the measurement matrix may contain many errors due to shadows and specular effects. Therefore, while in theory UPS can be solved with as few as three images, SVD can properly handle these modeling errors only when many images are supplied. Indeed, current methods \cite{alldrin,papa12} typically use 10 or more images. With fewer images SVD results tend to provide noisy solutions. Our method incorporates integrability into this estimation, providing valuable additional constraints that can reduce this noise. Our experiments indicate that our method can produce reasonable reconstructions with as few as 4 images and good reconstruction with 6 images, significantly improving over state-of-the-art methods with these few images.

For our approach we optimize a cost function based on \eqref{eq:ourcost} over the surface, lighting, normals, and (restored) error-free observations.  The cost ensures that normals and lighting are consistent with the measurements, which must have low rank.  We use constraints that ensure integrability.  This is somewhat tricky because rank constraints apply to the measurements while integrability constraints apply to the normals.  We show that by constructing a rank 3 matrix that contains normals, measurements, and lighting, we can impose the rank and integrability constraints together.  Specifically, we use a truncated nuclear norm approach \cite{tnn} to enforce the rank constraint, while integrability is represented by linear equalities.  This leads to a single non-convex problem  that we solve using a series of Alternate Direction Method of Multipliers (ADMM) operations \cite{boyd2011distributed,goldstein2014fast}.

Our formulation allows us to easily account for missing data in the measurement matrix.  This commonly occurs when pixels are dark due to shadows, or saturated due to specularities.  In some of the prior approaches, this can be solved with a preprocessing step, which may lead to a pipeline with yet another optimization \cite{wu}. We handle missing data using matrix completion based on the rank constraint. We initialize our optimization using prior approaches, in much the same way that bundle adjustment is initialized using simpler, but non-optimal algorithms \cite{Hartley2004}.



\section{Background and Previous Work}
\label{sec:back}

In this section we introduce in detail the problem of uncalibrated photometric stereo for Lambertian objects and review past work. We assume that we view an object in multiple images from a fixed viewpoint. In each image the object is illuminated by a single, distant point light source.  We represent lighting in image $i$ with $l_i \in R^3$, in which the direction of $l_i$ represents the direction to the lighting, and $\| l_i\|$ represents its magnitude.  We represent the object using a set of surface normals $\hat{n}_j \in R^{3}$, and albedos $\rho_j \in R$ for each pixel.   We then obtain images with the equation:
\begin{equation}
M_{ij} = \max(0, \rho_{j} l_{i}^T \hat n_j)
\label{eq:basic}
\end{equation}
where $M_{ij}$ represents the $j$-th pixel of the $i$-th image. We define the surface normal $\hat n_j=\frac{n_j}{\|n_j\|}$, $n_j = (-z_{x}, -z_{y},1)^{T}$, where $z_x$ and $z_y$ represent partial derivatives of the surface $z(x,y)$ at pixel $j$. Negative values of $\rho_{j} l_{i}^T \hat n_j$ are set to 0; these appear as attached shadows.

We now describe the creation of all images using matrix operations.  We define $S$ to be a $3 \times p$ matrix in which column $j$ contains $\rho_j\hat n_j$.  Given $m$ images, we can stack the light into the matrix $L$ of dimension $m \times 3$, where each row denotes one light per image. We concatenate all the images to form an observation matrix $M$ of dimension $m \times p$, where $p$ is the number of pixels. Now, in the absence of shadows, we can write the equation of UPS as:
\begin{equation} \label{eq:ps}
M = LS.
\end{equation}

Classical work on photometric stereo (e.g. \cite{woodham1980photometric}, see a recent review in~\cite{ackermann2015survey}) has assumed that known lighting is obtained by careful calibration. With $L$ known, \eqref{eq:ps} can be solved as a linear least squares problem. A more general and challenging case is unconstrained photometric stereo, in which the $L$ is unknown. A common approach, which we use as a baseline algorithm, follows the steps in Algorithm~\ref{algo:base}.

\begin{algorithm}[!h]
        \caption{Baseline}
        \label{algo:base}
        \begin{algorithmic}
                \STATE \textbf{Input}  : $M$
                \STATE \textbf{Output} : $Z$
                \STATE \textbf{Factorization} : Perform SVD on $M$ to obtain light and scaled surface normals $M=\tilde{L} \tilde{S}$ ~\cite{hayakawa}.
                \STATE \textbf{Integrability} : Follow Yuille and Snow ~\cite{yuille} to resolve ambiguity after the factorization using integrability. In $M=L S = \tilde{L} A^{-1} A \tilde{S}$, we solve for $A$, such that $S = A \tilde{S}$ approximately forms a set of integrable surface normals.
                \STATE \textbf{Depth Reconstruction} : Obtain the depth map $Z$ from the set of integrable surface normals $S$ as, e.g, in~\cite{basri}.
        \end{algorithmic}
\end{algorithm}

Belheumer \textit{et al.}~\cite{gbr} showed that in UPS the integrable set of surface normals can only be recovered up to a Generalized Bas-Relief transformation (GBR). A number of recent papers have concentrated on methods of solving the GBR ambiguity. Researchers have used priors on the albedo distribution~\cite{alldrin}, reflectance extrema~\cite{papa12}, grouping based on image appearance and color~\cite{self}, inter-reflections~\cite{reflec}, isotropy and symmetries~\cite{iso}, and specularity~\cite{spec} as constraints while solving for the GBR. All of these methods have first used the above mentioned baseline described in Algorithm \ref{algo:base} to obtain a solution up to the GBR.

Recent works have explored a variety of other research directions in photometric stereo. Mecca \textit{et al.}~\cite{mecca} proposed an integrated, PDE based approach to calibrated photometric stereo that uses a mere two images under perspective projection. It is not clear how to extend this to
uncalibrated photometric stereo. Basri \textit{et al.}~\cite{basri} extended the baseline to handle multiple light sources in each image using a spherical harmonics formulation. Chandraker \textit{et al.}~\cite{manu} proposed a method to handle attached and cast shadows in the case of multiple light sources per image. In \cite{sunkavalli2010visibility} the authors determine the visibility subspace for a set of images to remove the cast and attached shadows for performing UPS. Various works have addressed non-Lambertian materials (e.g., Georghiades \textit{et al.}~\cite{torrence} and Okabe \textit{et al.}~\cite{attach}).

In the context of Lambertian UPS, Georghiades \textit{et al.}~\cite{torrence} proposed to remove shadows and specularities and recover the missing pixel values using matrix completion algorithms, e.g., using the damped Wiberg~\cite{wiberg} or Cabral's algorithm~\cite{cabral}. Wu \textit{et al.}~\cite{wu} proposed a Robust PCA formulation as preprocessing for calibrated photometric stereo. Their approach seeks a low-rank (not necessarily rank 3) approximation to $M$ while removing outlier pixels (corresponding to shadows and specularities).  Oh \textit{et al.}~\cite{oh} applied Robust PCA in the context of calibrated photometric Stereo, replacing the Nuclear Norm with a Truncated Nuclear Norm (TNN) regularizer~\cite{tnn}. In \cite{papa12}, Favaro \textit{et al.} have used Robust PCA as preprocessing to the baseline algorithm for UPS.

\section{Our Approach}
\label{sec:approach}

In this section we introduce our integrated formulation that enforces integrability of surface normals in solving the uncalibrated  photometric stereo problem. We recall from~\eqref{eq:ps} that the measurement matrix $M$ can be factored into $M=LS$. To access the derivatives of $z(x,y)$ we write $S$ as a product
\begin{equation}  \label{equ:new_photo}
S = N \Lambda,
\end{equation}
where $N$ is a $3 \times p$ matrix whose $j$'th column is $n_j = (-z_{x}, -z_{y},1)^{T}$ and $\Lambda = \mathrm{diag}(\lambda_i, \lambda_2, ..., \lambda_p)$ with $\lambda_j = -\rho_j / \|n_j\|$. We next define the matrix:
\begin{equation}
X= \begin{bmatrix}
X^{I} & X^{N} \\
X^{L} & X^{M}
\end{bmatrix} =
\begin{bmatrix}
I \quad & N\\
L \quad & ~~M \Lambda^{-1}
\end{bmatrix},
\end{equation}
where $X$ is $(3+m)\times(3+p)$. The matrices $X$, $\Lambda,$ and the depth values ($z(x,y)$) form the unknowns in our optimization. Note that, because $LN=M \Lambda^{-1}$, the following holds for any $3 \times 3$ non-degenerate matrix $A$
\begin{equation}
X =   \begin{bmatrix}
A^{-1} \\ L A^{-1}
\end{bmatrix}\begin{bmatrix}
A & ~A N
\end{bmatrix}.
\end{equation}
This shows that $X$ is rank 3. The matrix $A$ represents a linear ambiguity. However, forcing the normals in $N$ to be integrable will reduce this ambiguity to the GBR.

To force integrability we denote by $\z=(z_1,...,z_p)^T$ the vector of unknown depth values and require
\begin{equation}  \label{eq:inte}
X^N = \begin{bmatrix} D_x \z,& D_y \z, & -\mathbf{1} \end{bmatrix}^T,
\end{equation}
where $D_x,D_y$ denote respectively the $x$- and $y$-derivative operators and $\mathbf{1}$ denotes the vector of all 1's.

Additional constraints are obtained by noticing that, because $0 \leq \rho_j \leq 1$ and $\|n_j\| \geq 1$,
\begin{equation}  \label{eq:lam_const}
-1 \leq \lambda_{j} \leq 0
\end{equation}
and
\begin{equation}  \label{eq:identity}
X^{I} = I_{3\times 3}.
\end{equation}

We are now ready to define our optimization function. Let $W$ be a binary, $m \times p$ matrix so that $W_{ij}=0$ if $M_{ij}$ is missing and $W_{ij}=1$ otherwise, and let
\begin{equation}  \label{eq:fdata}
f_{data}(X, \Lambda) = \frac{1}{2}||W \odot(M - X^{M} \Lambda)||^{2}_F,
\end{equation}
where $\odot$ denotes element-wise multiplication. Then~\eqref{eq:ourcost} can be written as
\begin{align}
\underset{X,\Lambda,\z} {\min} &\quad f_{data}(X,\Lambda) \nonumber \\
\mathrm{s.t.} & \quad \mathrm{rank}(X)=3, ~\eqref{eq:inte},
~\eqref{eq:lam_const}, \mathrm{and}~\eqref{eq:identity}.
\end{align}


\noindent
\textbf{Handling the rank-3 constraint:} Enforcing the non-convex constraint $\mathrm{rank}(X)=3$ can be challenging. In the context of matrix completion a recent paper~\cite{tnn} proposed using the Truncated Nuclear Norm (TNN) regularization term:
\begin{equation}  \label{eq:ftnn}
f_{tnn}(X) =  ||X||_* - \sum \limits_{k=1}^{3} \sigma_{k}(X),
\end{equation}
where $||X||_*$ denotes the nuclear norm of $X$ and $\sigma_k(X)$ is the $k$-th largest singular value of $X$. Clearly, $f_{tnn}(X)=0$ if and only if $\mathrm{rank}(X) \le 3$. We use $f_{tnn}$ as a regularizer and solve
\begin{align}  \label{eq:new_tnn}
\underset{X,\Lambda,\z} {\min} & \quad ~ f_{data}(X,\Lambda)  +  c ~ f_{tnn}(X) \nonumber \\
\mathrm{s.t.} & \quad~~ ~\eqref{eq:inte},
~\eqref{eq:lam_const}, \mathrm{and}~\eqref{eq:identity},
\end{align}
where $c$ is a preset scalar.



\section{Optimization using ADMM}
\label{sec:admm}

In this section we introduce a method for solving~\eqref{eq:new_tnn}. This is a challenging problem because both $f_{data}$ and $f_{tnn}$ are non-convex. Specifically, $f_{data}$~\eqref{eq:fdata} is bilinear in $X$ and $\Lambda$, while $f_{tnn}$~\eqref{eq:ftnn} is a difference between two convex functions. Our solution is based on a nested iteration in which the outer loop uses majorization to decrease $f_{tnn}$ whereas the inner loop uses the scaled ADMM with alternation to decrease $f_{data}$.

\bigskip

\noindent\textbf{Outer loop}: Following~\cite{tnn} at each iteration of the outer loop we replace $f_{tnn}(X)$ with a majorizer. Specifically, at iteration $k$ let $X^{(k)}=U\Sigma V^T$ be the singular value decomposition of $X^{(k)}$, and let $U_3$ (and $V_3$) be the matrices containing the left (right) singular vectors corresponding to the three largest singular values of $X^{(k)}$.  $U_3$ and $V_3$ are determined in the outer loop and are held constant throughout the inner loop. We then define
\begin{equation}
f_{maj}^{(k)}(X) = \|X\|_* - \trace(U_3^T X V_3).
\end{equation}
It was shown in~\cite{tnn} that $f_{maj}^{(k)}(X) \ge f_{tnn}(X)$ for all $X$ and that $f_{maj}^{(k)}(X^{(k)})=f_{tnn}(X^{(k)})$, and so decreasing $f_{maj}$ leads to decreasing $f_{tnn}$.

\bigskip

\noindent\textbf{Inner loop}:
In the inner loop we seek to minimize 
\begin{small}
\begin{align}
\underset{X,\Lambda,\z}{\min} \quad  &f_{data}(X,\Lambda) +  c f_{maj}^{(k)}(X)  \nonumber \\
\mathrm{s.t.} & \quad~~ ~\eqref{eq:inte}, ~\eqref{eq:lam_const}, \mathrm{and}~\eqref{eq:identity},
\label{eq:optimization}
\end{align}
\end{small}We use scaled ADMM, a variant of the augmented Lagrangian method that splits the objective function and aims to solve the different subproblems separately. We maintain a second copy of $X$, which we denote by $Y$ and form the augmented Lagrangian of~\eqref{eq:optimization} as follows
\begin{small}{\begin{align}
               \max_\Gamma \min_{X,\Lambda,\z,Y} & \quad \frac{1}{2}||W\odot(M - X^{M} \Lambda )||_F^2 + \nonumber \\ & c \left(||Y||_* - \trace(U_3^T Y V_3)\right)+ \frac{\tau}{2}||Y-X+\Gamma||_F^2  \nonumber \\
               s.t.  \ X^{I} = I_{3\times 3}, & ~
               -1 \leq \lambda_j \leq 0 ~ \forall j, ~
               X^N = \begin{bmatrix} D_x \z,& D_y \z, & -\mathbf{1} \end{bmatrix}^T,
               \label{eq:admm}
\end{align}}\end{small}where $||Y-X+\Gamma||^2_F$, denotes the Lagrangian penalty; $\tau$ is a constant, and $\Gamma$ is a matrix of Lagrange multipliers the same size as $X$ that is updated by the ADMM steps \cite{boyd2011distributed,goldstein2014fast}. We next describe the ADMM steps (applied iteratively).

\bigskip

\noindent
\textbf{\uline{Step 1}: Solving for $(X,\Lambda,\z)$}.\\
In each iteration, $k$, we solve the following sub-problems:
\begin{enumerate}
        \item Optimize w.r.t. $X^{I}$: $X^{I\,(k+1)} = I_{3 \times 3}$.
        \item Optimize w.r.t. $X^{L}$:
        \begin{eqnarray}
        X^{L\,(k+1)} &=& \underset{X^{L}}{\argmin}         ||Y^{L\, (k)}-X^{L}+\Gamma^{L\, (k)}||_F^2 \nonumber \\
        &=& Y^{L\,(k)} + \Gamma^{L\,(k)}.
        \label{eq:xL}
        \end{eqnarray}

        \item Optimize w.r.t. $X^{N}$ and $\z$:
        \begin{small}
        \begin{align}
        (X^{N\,(k+1)},\z^{(k+1)}) &= \underset{X^{N},\z}{\argmin}
        || Y^{N\,(k)}-X^{N}+\Gamma^{N\,(k)} ||_F^2 \nonumber  \\
        \mathrm{s.t.}~~ X^N &= \begin{bmatrix} D_x \z,& D_y \z, & -\mathbf{1} \end{bmatrix}^T.
        \label{eq:xN}
        \end{align}
        \end{small}The problem is solved by setting the third row of $X^{N\,(k+1)}$ to $-\mathbf{1}$ and by substituting $D_x\z$ and $D_y\z$ for the first two rows of $X^N$ in the objective, obtaining linear least squares equations in $\z$ that can be solved directly.

        \item Optimize w.r.t. $X^{M}$ and $\Lambda$:
        \begin{small}
        \begin{align*}
        (X^{M\,(k+1)},\Lambda^{(k+1)} )= &
        \underset{X^{M},\Lambda}{\argmin}
        \frac{1}{2}\|W\odot(M - X^{M}\Lambda )\|_F^2  \\
        & +\frac{\tau}{2} \|Y^{M\,(k)}-X^{M}+\Gamma^{M\,(k)}\|_F^2 \nonumber \\
        ~~~ \mathrm{s.t.} ~~~  -1 \leq \lambda_j \leq 0  ~ \forall j.
        \label{eq:new_lam}
        \end{align*}
        \end{small}We will separate this into the known and unknown pixels based on $W$.
        For an \textbf{unknown pixel} $j$ in frame $i$ ($W_{ij}=0$) the first term vanishes and the minimization only determines the respective entry of $X^{M}$ so that:
        \begin{equation}
        X^{M\, (k+1)}_{ij} = Y_{ij}^{M\,(k)}+\Gamma_{ij}^{M\,(k)}.
        \label{eq:XM_no}
        \end{equation}

        For the \textbf{known pixels}, since $\Lambda$ is diagonal we can write these equations separately for each column $j$ (corresponding to the $j$-th pixel):
        \begin{small}
        \begin{align}
        (X^{M\,(k+1)}_{j},\lambda_{j}^{(k+1)}) = &
        \underset{X_{j}^{M},\lambda_{j}}{\argmin}
        \frac{1}{2} \| (W_j \odot (M_j - \lambda_{j}
        X_{j}^{M} )\|^{2}_{2}  \nonumber \\
        &+ \frac{\tau}{2}\|Y_{j}^{M\,(k)}-X_{j}^{M}+\Gamma_{j}^{M\,(k)}\|_{2}^{2} \nonumber \\
        ~~~ \mathrm{s.t.} ~~ -1 \leq \lambda_j \leq 0.
        \label{eq:new_lam_p11}
        \end{align}
        \end{small}The problem \eqref{eq:new_lam_p11} is non-convex. We will solve it with \textit{alternate optimization}. $X^{M}$ and $\Lambda$ are updated by the following steps until convergence.

        $\mathbf{X^{M}}$ : Let $\tilde{M_{j}}=W_{j} \odot M_{j}$ , $\tilde{X_{j}}=W_{j}      \odot X^{M}_{j}$ and $\tilde{A}^{M\,(k)}_{j}=W_{j} \odot (Y_{j}^{M\,(k)}+\Gamma_{j}^{M\,(k)})$.      Then,
                \begin{align}
                \tilde{X_{j}} &=
               \underset{\tilde{X_{j}}} \argmin ~ \frac{1}{2} || \tilde{M_j} - \lambda_{j} \tilde{X_j}||^{2}_{2} + \frac{\tau}{2} || \tilde{A}^{M\,(k)}_j - \tilde{X_j}||^{2}_{2}         \nonumber \\
                &= \dfrac{\lambda_j \tilde{M_j} + \tau \tilde{A}^{M\,(k)}_j}{\lambda^{2}_j + \tau}.
                \label{eq:xM}
                \end{align}
         $\mathbf{\Lambda}$: \begin{align}
					\lambda_j &= \underset{\lambda_j} \argmin ~
					\frac{1}{2} \| \tilde{M_j} - \lambda_j \tilde{X_j}||^2_2 ~~ \mathrm{s.t.} ~ -1 \leq \lambda_j \leq 0, \nonumber \\
					&=\min ( 0,\max ( -1, \tilde X_j^T \tilde M_j/ \| \tilde X_j \|^2_2 ) ).
					\label{eq:l}
	               \end{align}
	               \end{enumerate}

\noindent
\textbf{\uline{Step 2}: Solving for $Y$.}
Solving for $Y$ requires a solution to
\begin{eqnarray}  \label{eq:solvey}
Y^{(k+1)} = \argmin_{Y} c \left(\|Y\|_{*} -  \trace(U_3^T Y V_3) \right) \nonumber \\
+ \frac{\tau}{2} ||Y - X^{(k+1)} + \Gamma^{(k)}||_F^2.
\end{eqnarray}
Below we show that this problem can be solved in closed form by applying the shrinkage operator, obtaining
\begin{equation}
Y^{(k+1)} = D_{\nicefrac{c}{\tau}}(X^{(k+1)} - \Gamma^{(k)}+\dfrac{c}{\tau}U_3V_3^T),
\label{eq:Y}
\end{equation}
where the shrinkage operator $D_{t}(.)$ is defined as follows. For a scalar $s$ we define $D_t(s)=\mathrm{sign}(s)\times\max(|s|-t,0)$. For a diagonal matrix $S=\mathrm{diag}(s_1,s_2,...)$ with non-negative entries we define $D_t(S)=\mathrm{diag}(D_t(s_1),D_t(s_2),...)$. Finally, for a general matrix $\Upsilon$, let $\Upsilon=\tilde U S \tilde V^T$ be its singular value decomposition, then $D_{t}(\Upsilon)=\tilde U D_t(S) \tilde V^{T}$.\\

To derive~\eqref{eq:Y}, we rewrite~\eqref{eq:solvey} as:
\begin{small}
\begin{equation}
Y^{(k+1)} = \argmin_{Y} \|Y\|_{*} +
\dfrac{\tau}{2 c} \| Y - X^{(k+1)} + \Gamma^{(k)} -
\dfrac{c}{\tau}U_3 V_3^{T} \|^{2}_{F} - T,
\label{eq:sim_y}
\end{equation}
\end{small}where $T=\trace( V_3U_3^T (X^{(k+1)}-\Gamma^{(k)}))+\dfrac{c}{2 \tau}\| U_3V_3^T\|_F^2$ is independent of $Y$. Equation \eqref{eq:sim_y}
is of the general form $\underset{Y} \min \|Y\|_{*} + \dfrac{1}{2t}\|Y -C\|^2_F$, for which the solution is $D_t(C)$, as is shown in~\cite{cai2010shrinkage}, implying~\eqref{eq:Y}.


\noindent
\textbf{\uline{Step 3}: Update of $\Gamma$.}
The matrix $\Gamma$ contains Lagrange multipliers that are used in the saddle-point formulation \eqref{eq:admm} to enforce the equality constraint $X=Y$.    The following update is a gradient ascent step that acts to maximize the augmented Lagrangian  \eqref{eq:admm} for $\Gamma.$  For details, see \cite{boyd2011distributed,goldstein2014fast}.
\begin{eqnarray}
\Gamma^{(k+1)} = \Gamma^{(k)} + (Y^{(k+1)} - X^{(k+1)}).
\label{eq:G}
\end{eqnarray}

The entire optimization process is listed in Algorithm~\ref{algo:rtnn}. We will make the code available.
\begin{algorithm}[!h]
        \caption{TNN formulation solved with ADMM}
        \label{algo:rtnn}
        \begin{algorithmic}
                \STATE \textbf{Input:} $M$, $W$.
                \STATE \textbf{Output: } $X$, $\z$.
                \STATE \textbf{Initialization:} Initialize $X^{L}$ and $X^{N}$ by running Baseline algorithm (without resolving GBR). Initialize $X^{M} = - M$, $\Lambda=-I$, and $c=1$. Set $X^{(0)}=X$, $Y=X$, $\Gamma=0$, and $\tau=1$.
                \STATE{$k=0$.}
                \WHILE{not converged}
                \STATE Perform SVD over $X^{(k)}$ to obtain $U_3$ and $V_3$.
                \STATE \textbf{Run ADMM:}
                \WHILE{not converged}
                \STATE{\textbf{Update of $X$, $\z$ and $\Lambda$}}.
                \STATE{Update $X^{I(k+1)}=I_{3 \times 3}$}.
                \STATE{Update $X^{L(k+1)}$ using (\ref{eq:xL})}.
                \STATE{Update $X^{N(k+1)}$ and $\z$ using (\ref{eq:xN})}.

                \WHILE{not converged}
                
                \FOR{each pixel $j$}
                \STATE{Update $X^{M(k+1)}_{j}$ using (\ref{eq:xM}) and $\lambda_{j}^{(k+1)}$ using (\ref{eq:l})}.
                \ENDFOR
                \FOR{each pixel $j$ in each image $i$}
                \IF{$W_{ij}$ = 0 i.e. pixel $j$ is not known}
                \STATE{Update $X^{M(k+1)}_{ij}$ using (\ref{eq:XM_no})}.
                \ENDIF
                \ENDFOR
                \ENDWHILE
                \STATE{\textbf{Update $Y$} using (\ref{eq:Y})}.
                \STATE{\textbf{Update of $\Gamma$} using (\ref{eq:G})}.
                \STATE{$k=k+1$.}
                \ENDWHILE
                \ENDWHILE
        \end{algorithmic}
\end{algorithm}

\section{Experimental Results}
\label{sec:res}

In this section we evaluate and compare the performance of our algorithm with two versions of the baseline algorithm, in both real world and synthetic examples. We compare the following methods:\\
\textbf{Baseline}: Algorithm~\ref{algo:base} described in Section~\ref{sec:back}. This method is used in ~\cite{alldrin,papa12,self,reflec,iso,spec}.\\
\textbf{RPCA}: Images are preprocessed using Robust PCA \cite{wu}, parameters are chosen as suggested by~\cite{papa12}.  Then we apply the baseline algorithm to the obtained matrix.  This method is used in~\cite{papa12}. RPCA solves a sparse low rank optimization to detect shadows and other non-Lambertian effects. The method uses $L_1$ regularization to identify outlier pixels, even when they do not result in intensities near 0 or 1.  \\
\textbf{Our(NC)}: Our  proposed formulation as described in Section~\ref{sec:admm} using $W=1$, i.e., no completion. This allows comparison to Baseline, which also does not perform matrix completion.\\
\textbf{Our(MC)}: Our proposed formulation as in Section~\ref{sec:admm} with $w_{ij} \in \{0,1\}$, allowing for matrix completion. In both versions of our algorithm we use $c=1$ and use RPCA to initialize optimization. We identify missing pixels as those with normalized intensity outside the range of $(0.02,0.98)$.

All the tested methods solve for the surface only up to a GBR ambiguity.  To compare the results with ground truth, we find the GBR that optimizes the fit to ground truth, and measure the residual error.

In the presence of a large number of images with noise and non-Lambertian effects, we expect the sequential pipeline of Baseline and RPCA, involving SVD, to produce accurate solutions, because the problem solved by SVD is heavily overconstrained.  In the presence of fewer images, our integrated method will be able to produce a more accurate decomposition by using both rank and integrability constraints to find the right linear subspace. Thus we expect our integrated approach to improve over the Baseline and RPCA as we reduce the number of images. In the following sub-section we will show results with synthetic and real world data that supports our claim.

\subsection{Experiments on Synthetic Data}
\label{sub:syn}

We use five real objects (``cat'', ``owl'', ``rock'', ``horse'', ``buddha'') to produce synthetic images, their shape is obtained by applying calibrated photometric stereo to a publicly available dataset~\cite{data1}. We use the normals and albedos from these objects to generate images.  Each image is generated by a randomly selected light source which lies at 30 degrees of the viewing direction on average. All images are of size 512 $\times$ 340 with objects occupying 29-72K pixels. A segmentation mask is also supplied. To show the variation of performance with the number of images $N_I$, we use sets of 4, 6, 8, 10, 15, 20, 25 and 30 images respectively. We add Gaussian noise with standard deviation ranging from 1\% to 7\% (in steps of 2\%) of the maximum intensity.  For each choice of noise, we run 5 different trials with random noise and lighting to generate the synthetic images. Thus we have 5 objects, 4 levels of noise and 5 random simulations, making a total of 100 experiments for each of the 8 different sets of 4, 6, 8, 10, 15, 20, 25 and 30 images. As a measure of performance, we calculate the error in the reconstructed depth map. Let the ground truth surface be $Z_T$ and the reconstructed surface be $Z_{rec}$. We measure error in depth as $Z_{err}=100 \times \frac{||Z_T - Z_{rec} ||}{|| Z_T ||}$. To compare two algorithms (say, algorithm A vs.\ algorithm B), we define the following two terms :\\
\textbf{Relative Improvement (in $\%$)} : Denote $e^a_k$ and $e^b_k$ as the depth error for each trial $k$ by using algorithm A and B respectively. The Relative Improvement of algorithm B over A is the average of $\frac{(e^b_k - e^a_k)}{e^b_k}$ over all trials $K$ for each choice of $N_I$ expressed in percentage.\\
\textbf{Percent of Improved Trials} : This denotes the number of trials in which algorithm B improves over A. In terms of notation introduced previously, this is $\frac{1}{K}\sum_{k=1}^K \mathbb{I}(e^{\mathrm{a}}_k < e^{\mathrm{b}}_k)$, where $\mathbb{I}(.)$ is in indicator variable and $K$ is the total number of trials for each choice of $N_I$. The measure is expressed in percentage.

%
\begin{figure}[!h]
       \centering
\begin{subfigure}[b]{0.4\textwidth}
       \includegraphics[width=\textwidth]{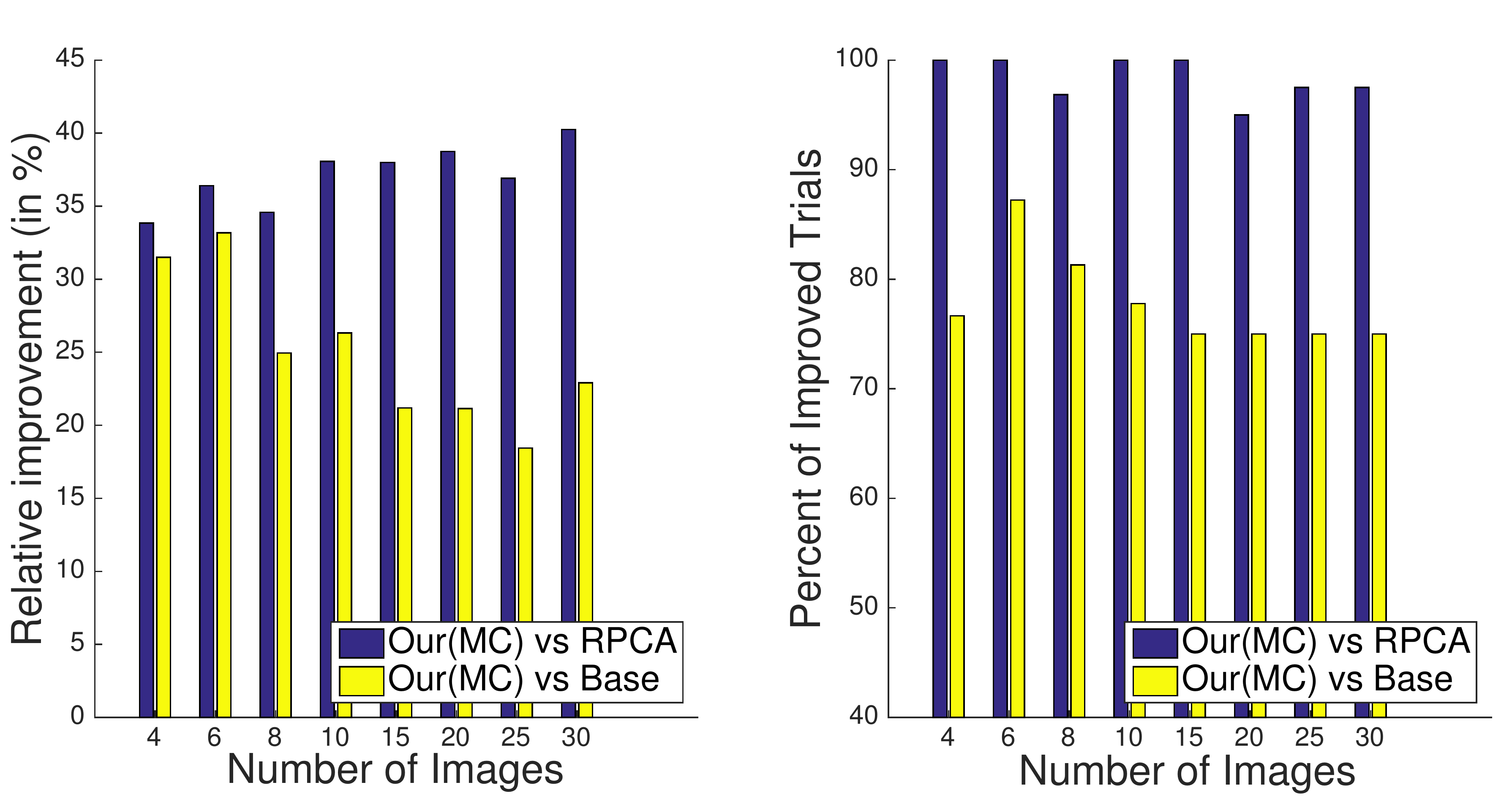}
       \caption{Gaussian noise}\label{fig:syn_rpca}
\end{subfigure}
\begin{subfigure}[b]{0.4\textwidth}
	\includegraphics[width=\textwidth]{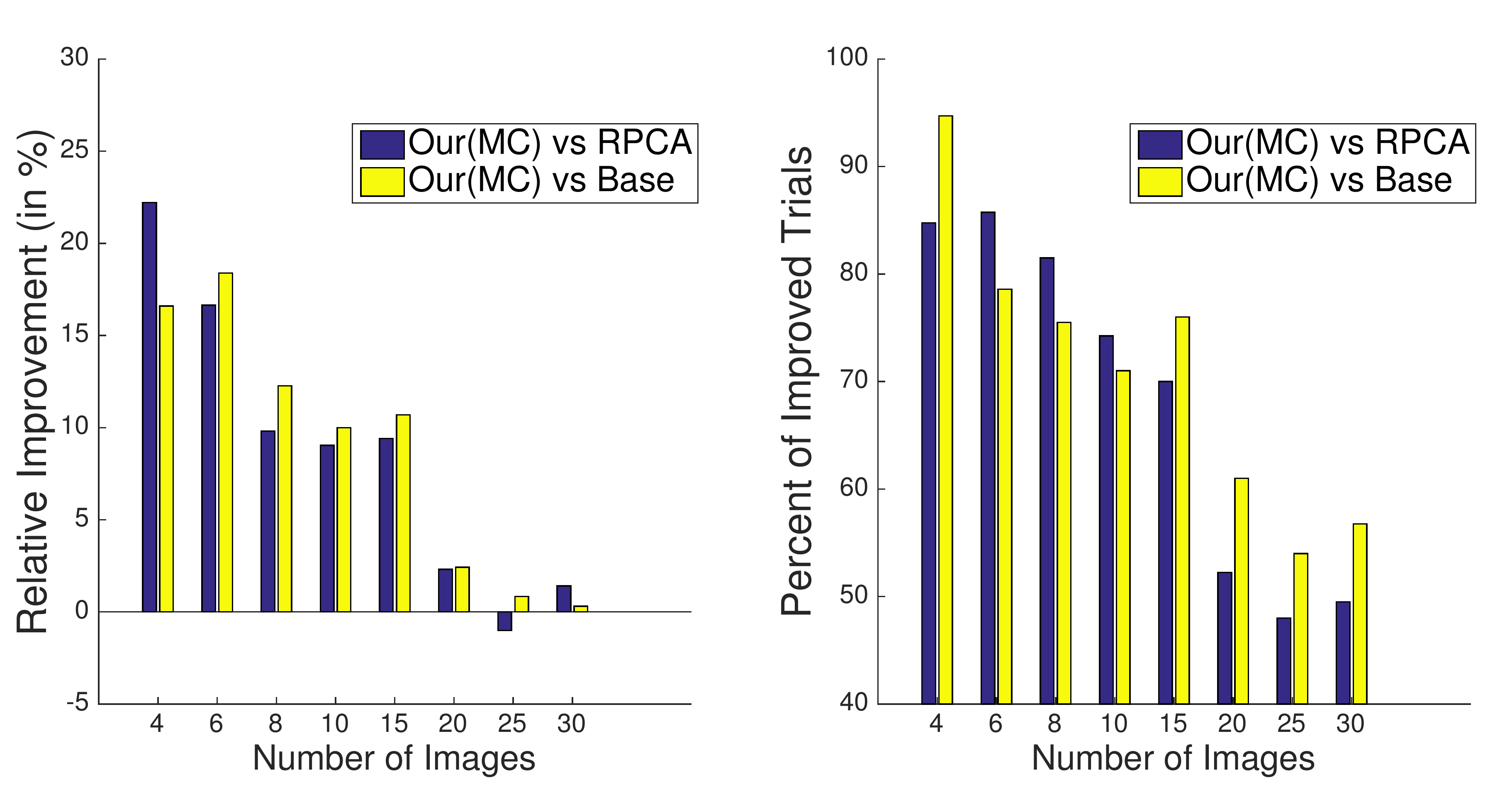}
	\caption{Gaussian noise with Phong model}\label{fig:img_ph}
\end{subfigure}

\caption{\small Performance comparison of Our(MC) algorithm to RPCA (in blue) and Baseline (yellow) for different numbers of input images with gaussian noise under either a pure lambertian model (top) or the Phong model (bottom). The left bar plot shows the amount of relative improvement achieved with our algorithm, and the right plot shows the percent of trials in which our algorithm out performed each one of the competing algorithms.}
\label{fig:syn_ph}
\end{figure}

\begin{figure}[!h]
	\centering
	\includegraphics[width=0.48\textwidth]{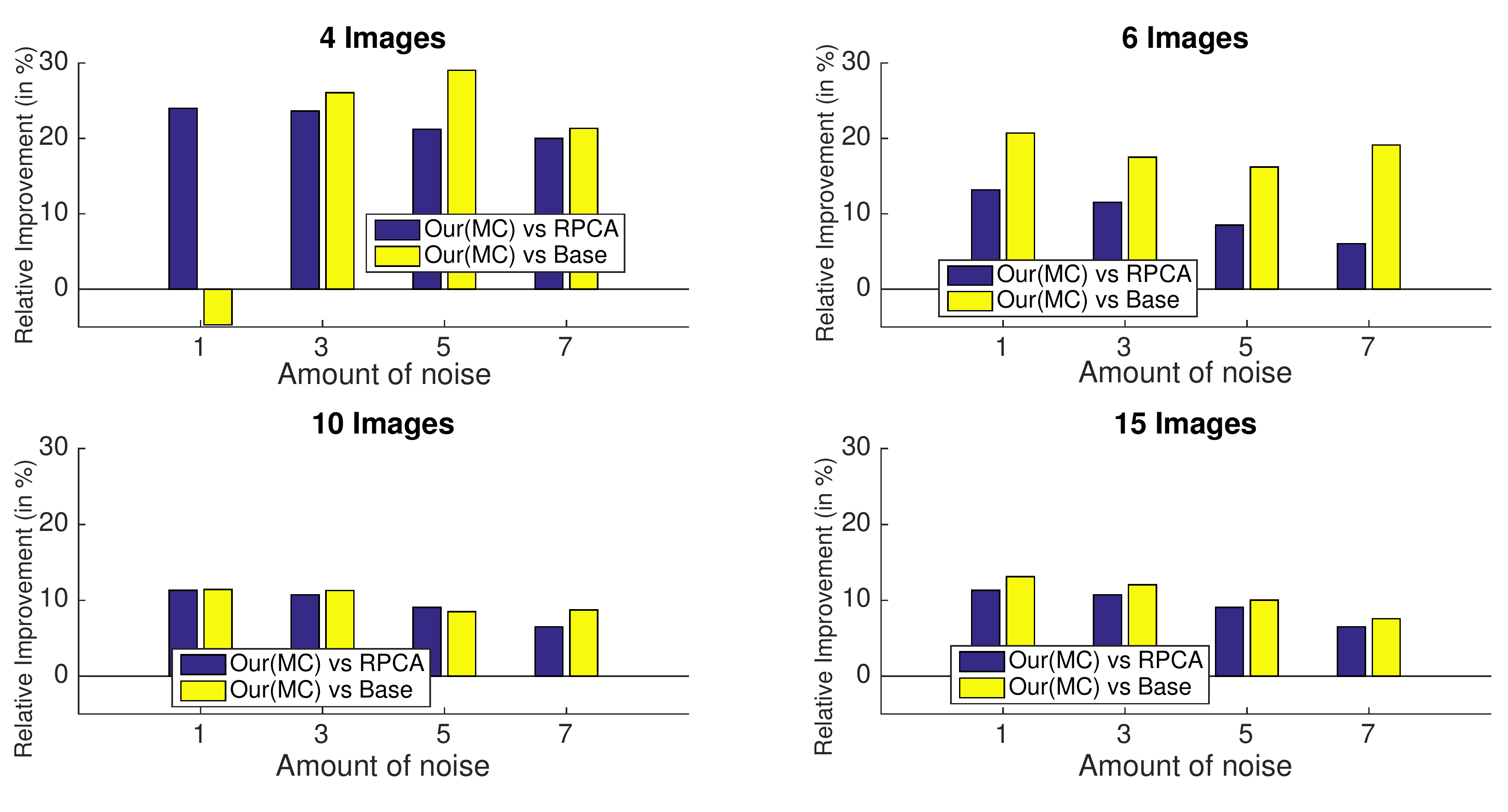}
	\caption{\small Performance comparison of  Our(MC) with RPCA and Baseline with varying noise created using the Phong model.}
	\label{fig:noise_ph}
\end{figure}

In Figure~\ref{fig:syn_rpca} we compare performance of Our(MC) with Baseline and RPCA, on synthetic data in the presence of Gaussian noise. We initialize our methods with RPCA. We observe that as  the number of images decreases, our method improves compared to Baseline and RPCA. With simple Gaussian noise RPCA doesn't produce additional advantages as there are no outliers.

In Figure~\ref{fig:img_ph} we compare the performance of our methods on synthetic data with Gaussian noise and with specularities generated by the Phong reflectance model~\cite{phong1,phong2}. Mathematically each image $M_{i}$ can be represented as :
\begin{equation}
M_{i} = L_{i}S + k_{s} (VR)^{\alpha},
\end{equation}
where $V$ is the viewing direction and $R$ denotes the directions of perfect reflection for incoming light $L_i$ for each pixel $j$.  Larger $\alpha$ produces sharper specularities, while larger $k_s$ causes more light to be reflected as specularity.  We use $k_s = 0.2$ and $\alpha = 10$. We observe that the advantage of Our(MC) degrades as the number of images increases,  as expected. This experiment shows that even though our method is designed specifically for Lambertian objects it can tolerate a certain amount of model irregularities such as specularity. With 4 images our method beats RPCA in 85\% of the all trials with a relative improvement of 22.12\%.

In Figure \ref{fig:noise_ph} we compare Our(MC) with Baseline and RPCA with variation of noise for different subsets of images (4,6,10 and 15). We can conclude that our method is robust to noise and its advantages do not degrade with an increase in noise.





\subsection{Experiments on Real World Data}
\label{sub:real}

\begin{figure}[!h]
	\centering
	\includegraphics[width=0.48\textwidth]{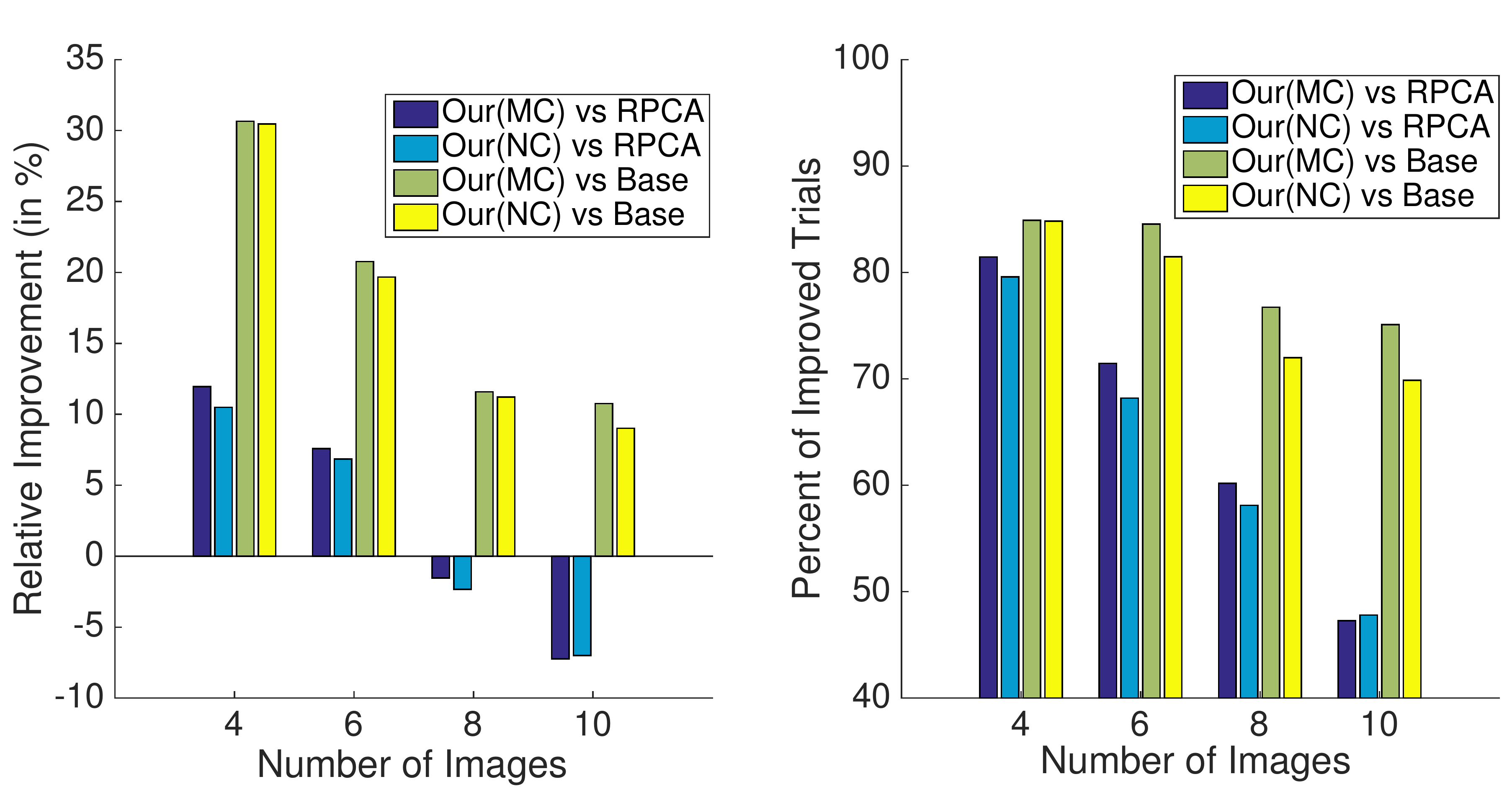}
	\caption{\small Performance comparison of Our (MC) and Our (NC) algorithms to RPCA and Baseline with real images.}
	\label{fig:real}
\end{figure}

To test our approach on real data, we used the two publicly available data sets~\cite{data1} and~\cite{data} consisting of 5 and 7 objects respectively.  The datasets provide calibrated lighting, which we use to perform calibrated photometric stereo. The obtained depth map, albedo, and surface normals are considered as ground-truth for photometric stereo with unknown lighting similar to \cite{alldrin}. To show the variation of performance with the number of images, we select subset of 4, 6, 8 and 10 images for each object. We perform 10 random selections of subset of images for each of the 12 objects. Thus we have 120 experiments for every subset of images.

In Figure \ref{fig:real} we compare the performance of our methods, Our(MC) and Our(NC), with Baseline and RPCA with variation in the number of images. We see that for fewer images our methods outperform Baseline and RPCA by a significant amount and are comparable to RPCA for more images. For 4 images Our(MC) outperforms Baseline in 84.9\% cases with a relative improvement of 30.6\% and outperforms RPCA in 81.4\% cases with a relative improvement of 12\%. However for 10 images we beat Baseline in 75\% cases with a relative improvement of 10.7\% and beat RPCA in only 47.3\% cases with a relative improvement of -7.2\%.

\begin{figure}[!h]
	\centering
	\includegraphics[width=0.45\textwidth]{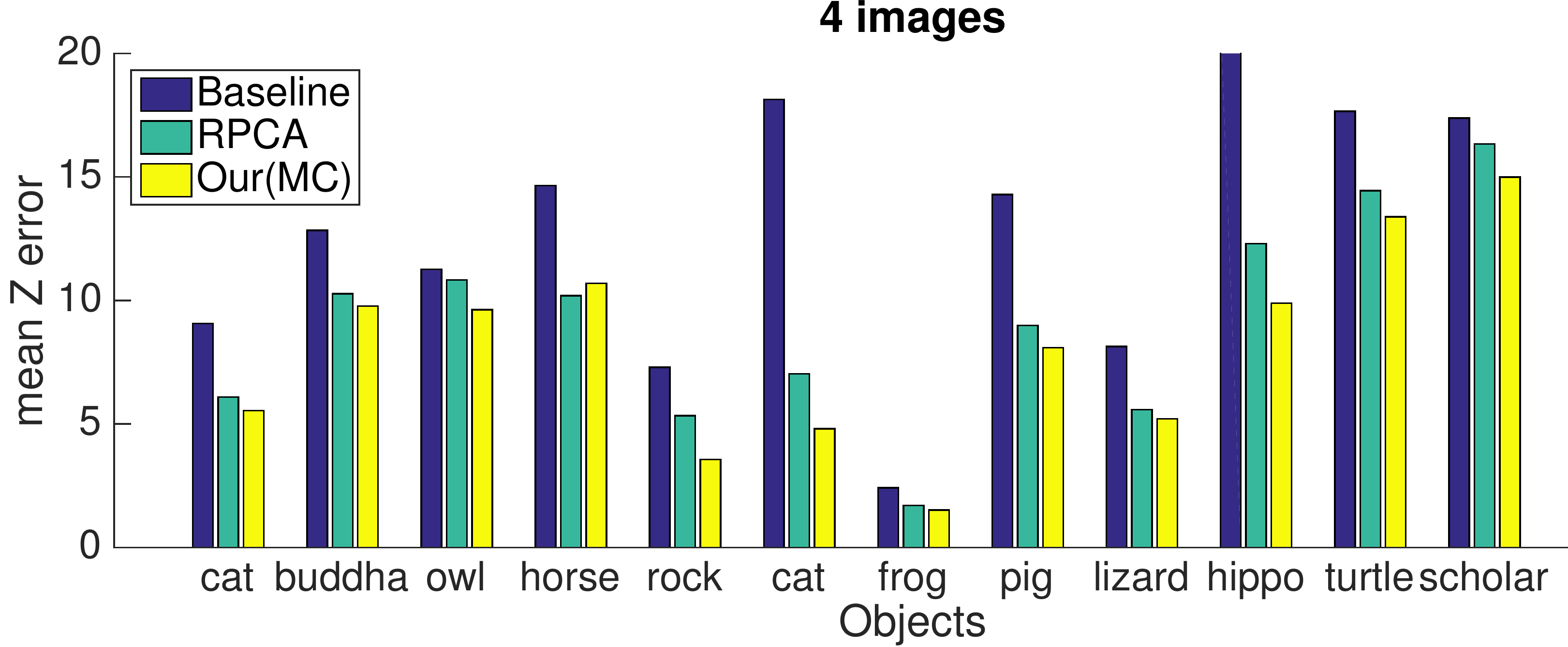}
	\vspace{2mm}
	\includegraphics[width=0.45\textwidth]{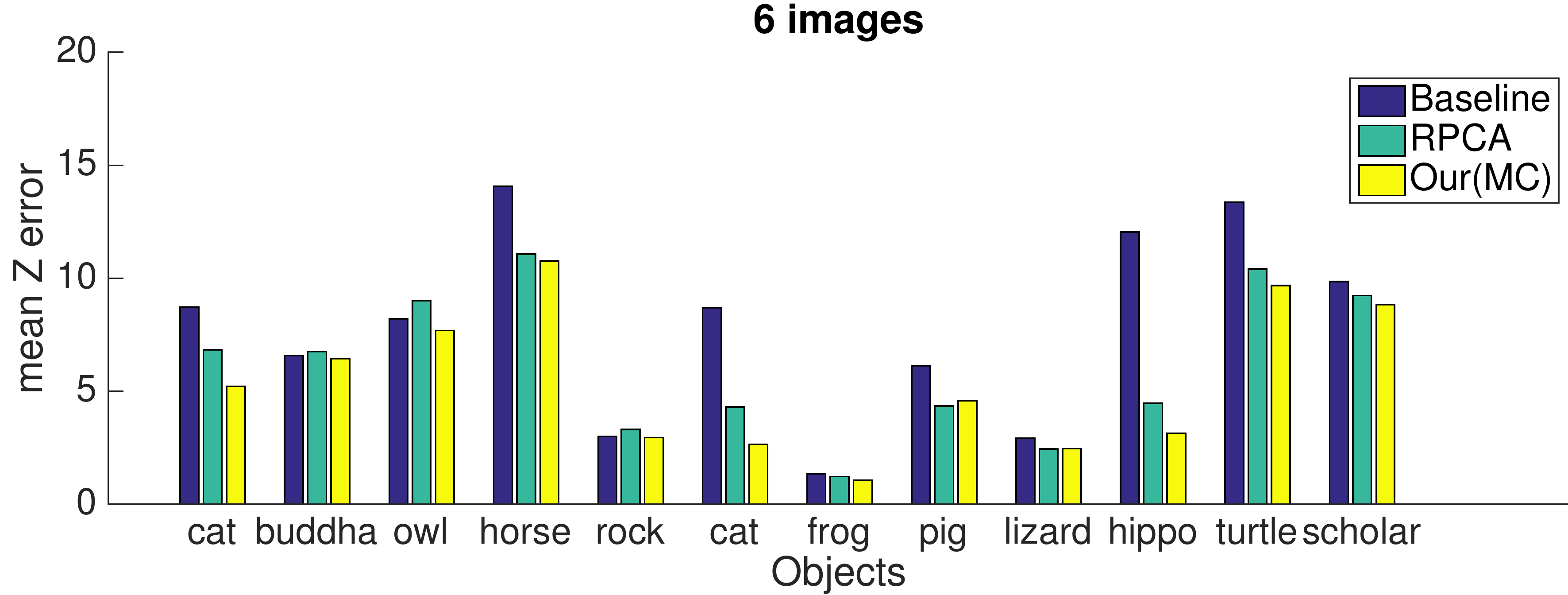}
	\caption{\small Average surface reconstruction error with 4 (top) and 6 (bottom) real images of 12 objects over 10 random trials using Our(MC), RPCA and Baseline.}
	\label{fig:im}
\end{figure}

\begin{figure*}[!ht]
	\centering
	\begin{subfigure}[b]{0.48\textwidth}
		\includegraphics[width=\textwidth]{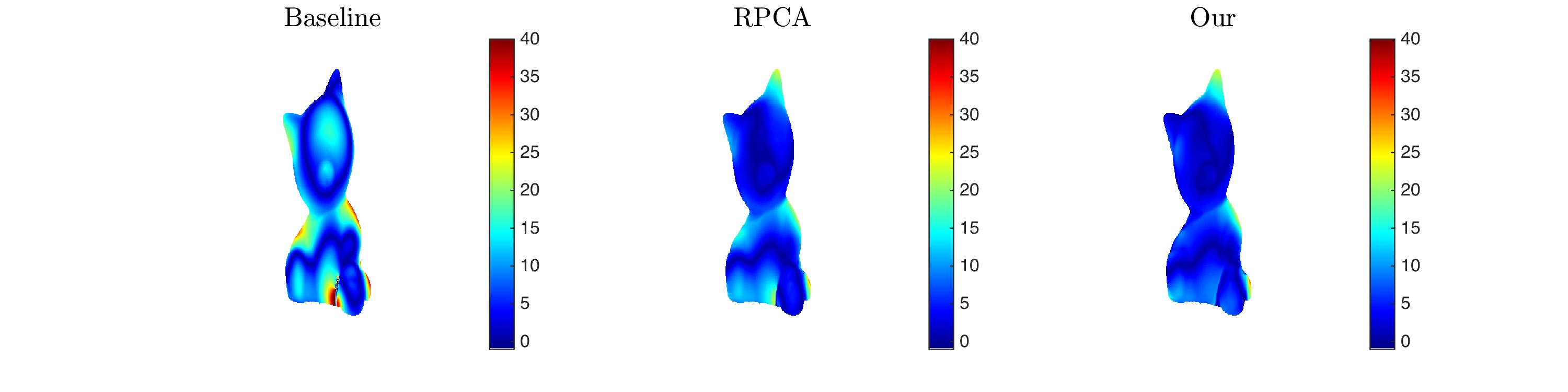}
	\end{subfigure}
	\begin{subfigure}[b]{0.48\textwidth}
		\includegraphics[width=\textwidth]{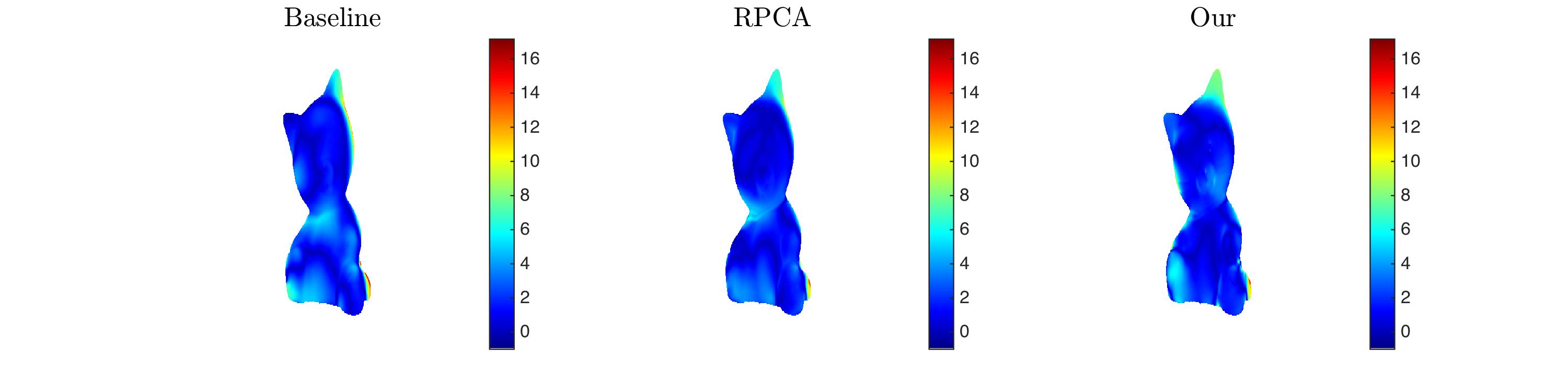}
	\end{subfigure}
	\begin{subfigure}[b]{0.48\textwidth}
		\includegraphics[width=\textwidth]{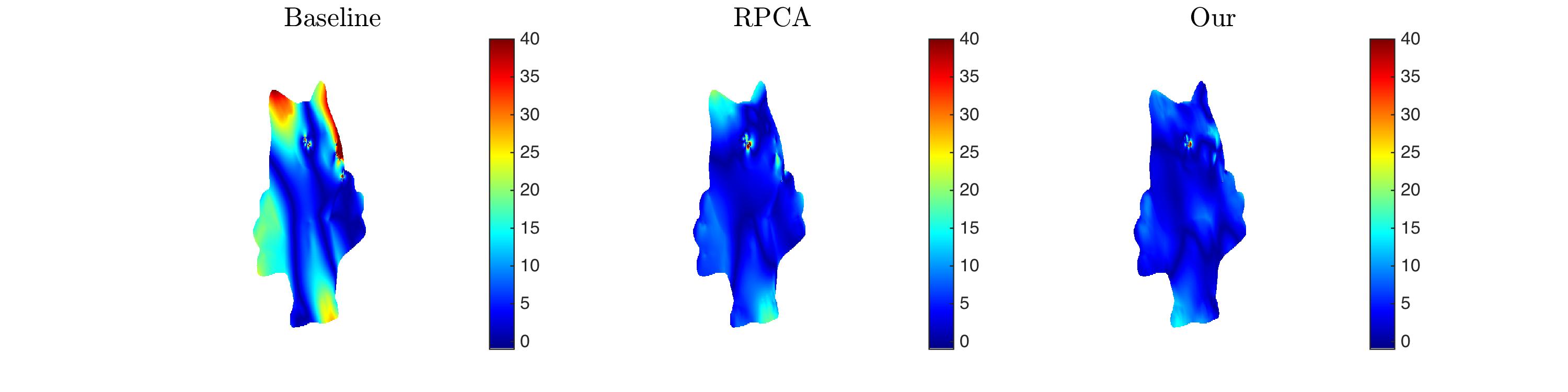}
	\end{subfigure}
	\begin{subfigure}[b]{0.48\textwidth}
		\includegraphics[width=\textwidth]{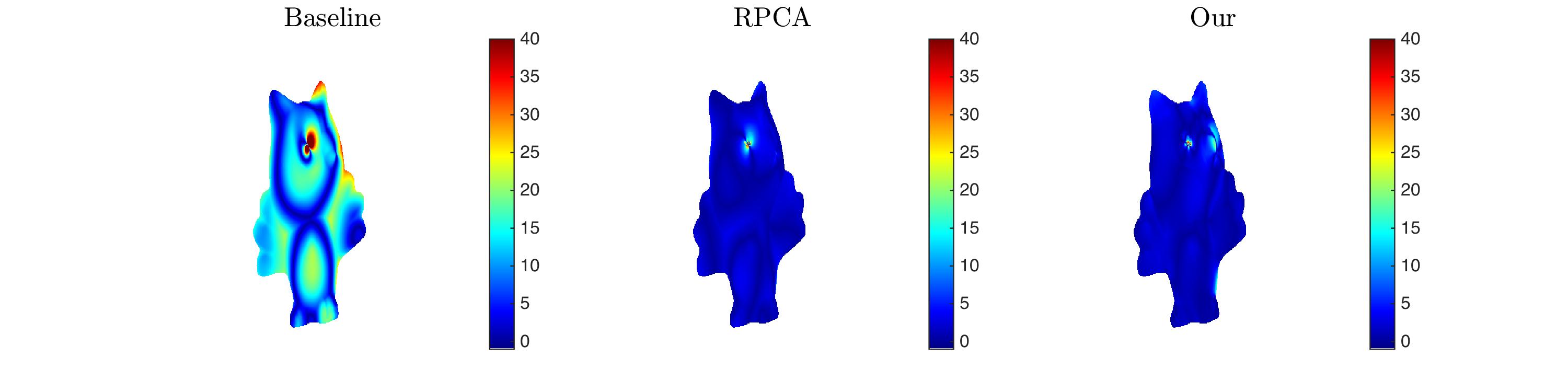}
	\end{subfigure}
	\\
	\begin{subfigure}[b]{0.48\textwidth}
		\includegraphics[width=\textwidth]{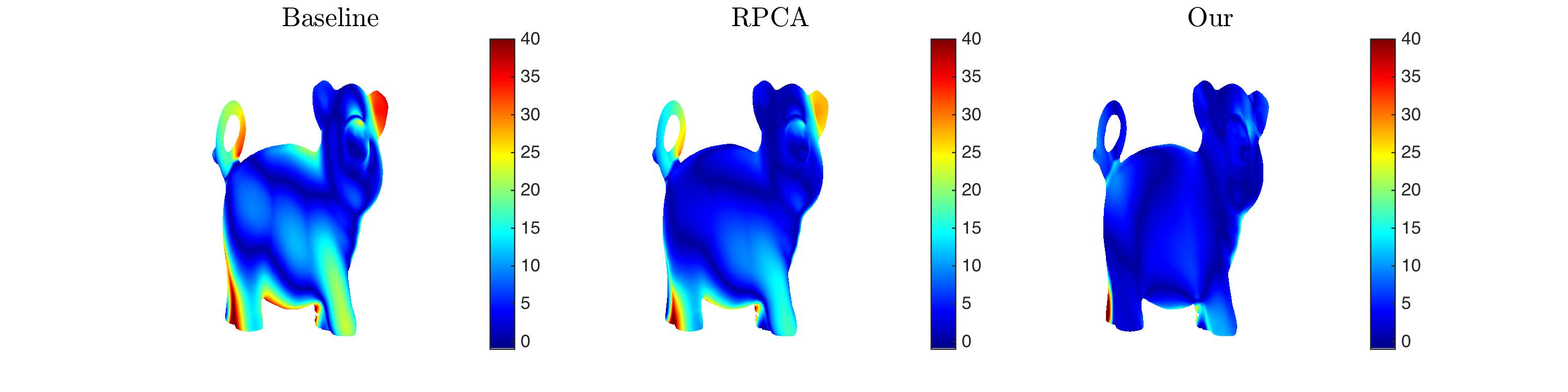}
	\end{subfigure}
	\begin{subfigure}[b]{0.48\textwidth}
		\includegraphics[width=\textwidth]{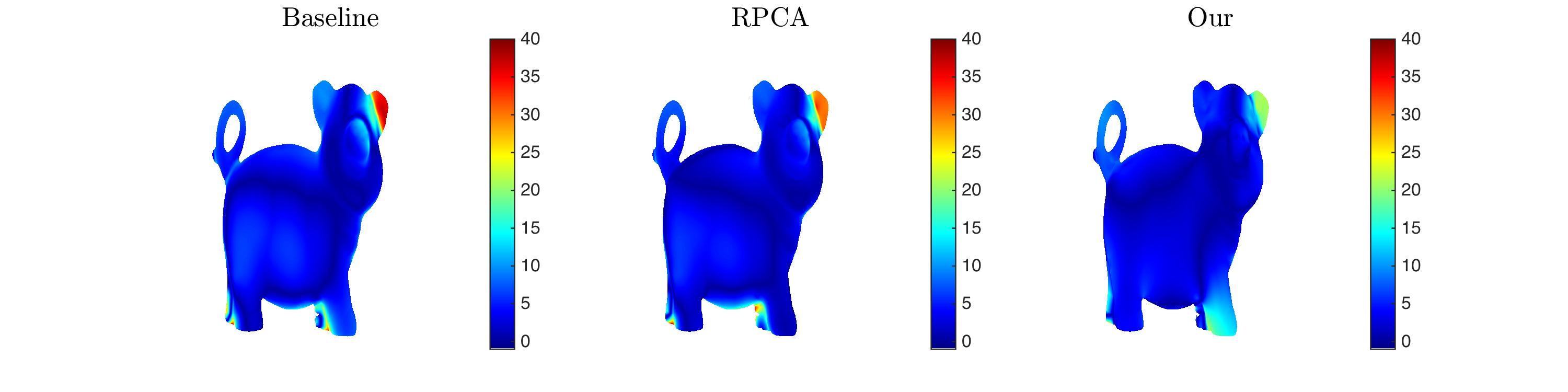}
	\end{subfigure}
	\begin{subfigure}[b]{0.48\textwidth}
		\includegraphics[width=\textwidth]{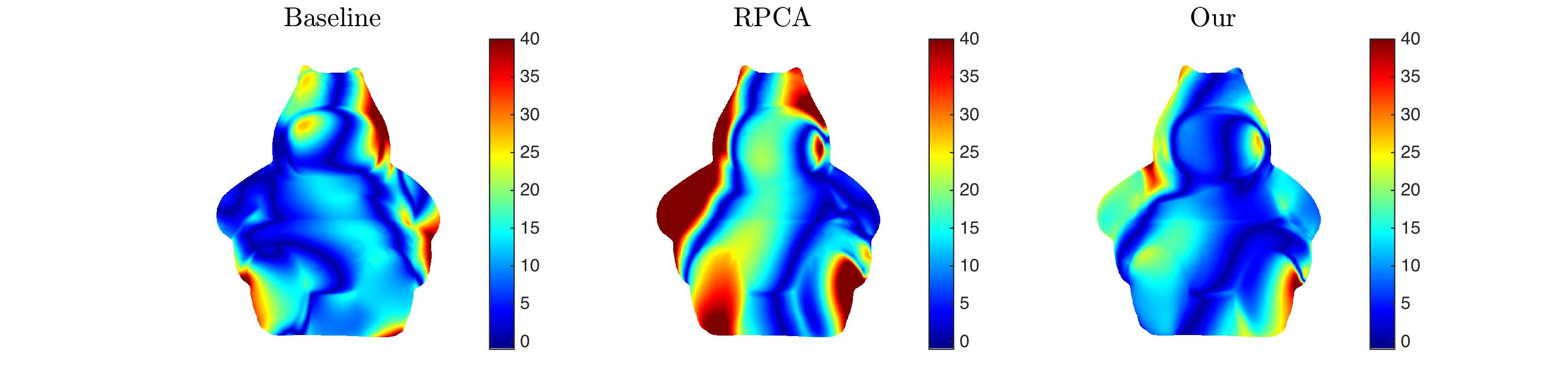}
	\end{subfigure}
	\begin{subfigure}[b]{0.48\textwidth}
		\includegraphics[width=\textwidth]{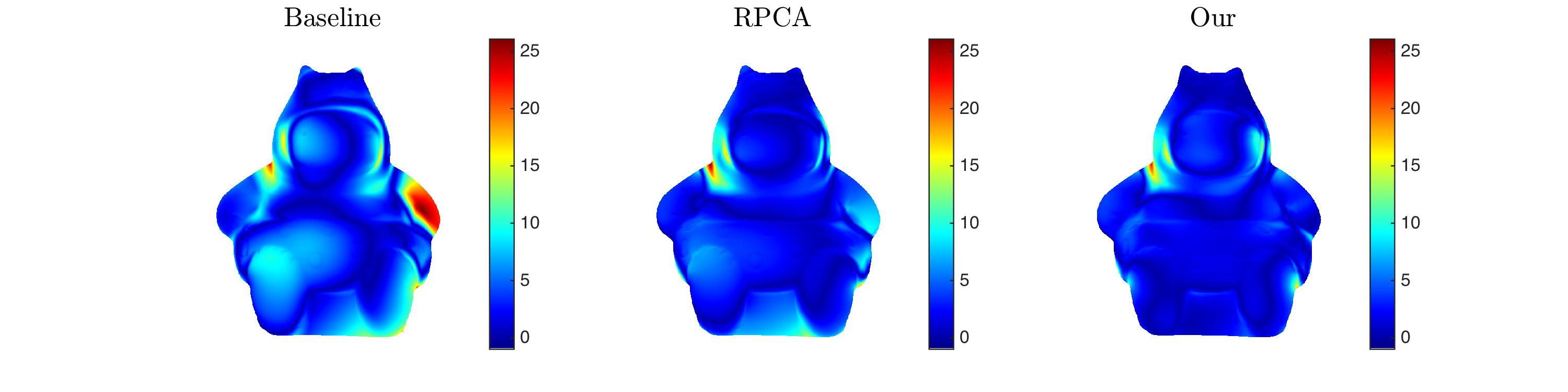}
	\end{subfigure}
\caption{\small Reconstruction error $|Z_T - Z_{rec}|$ for Baseline, RPCA and Our(MC) on ``Cat", ``Owl", ``Pig" and ``Hippo'' shown in each row. The left column shows results for 4 images, the right shows results for 10.}
	\label{fig:zerr_some}
\end{figure*}

\begin{figure*}[!ht]
	\centering
	\begin{subfigure}[b]{0.22\textwidth}
		\includegraphics[width=\textwidth]{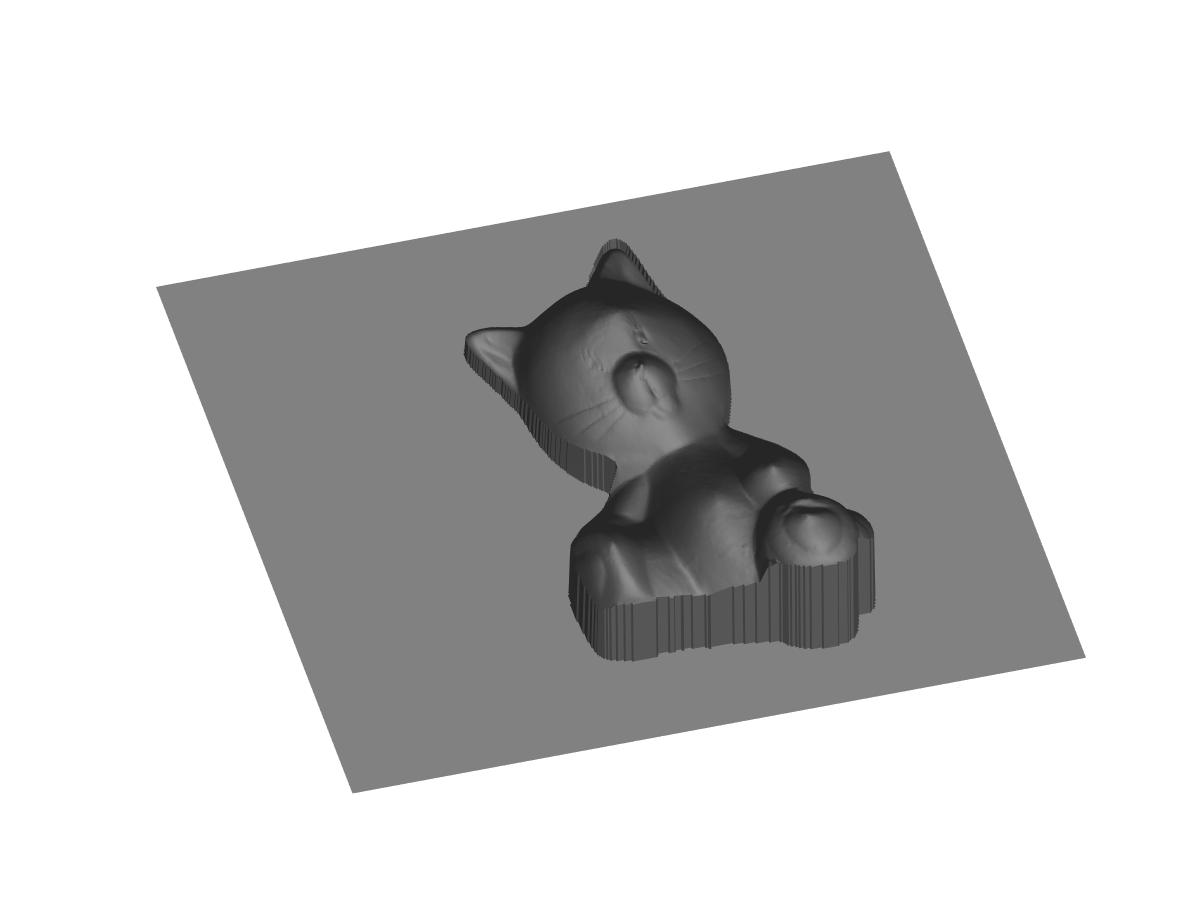}
		\includegraphics[width=\textwidth]{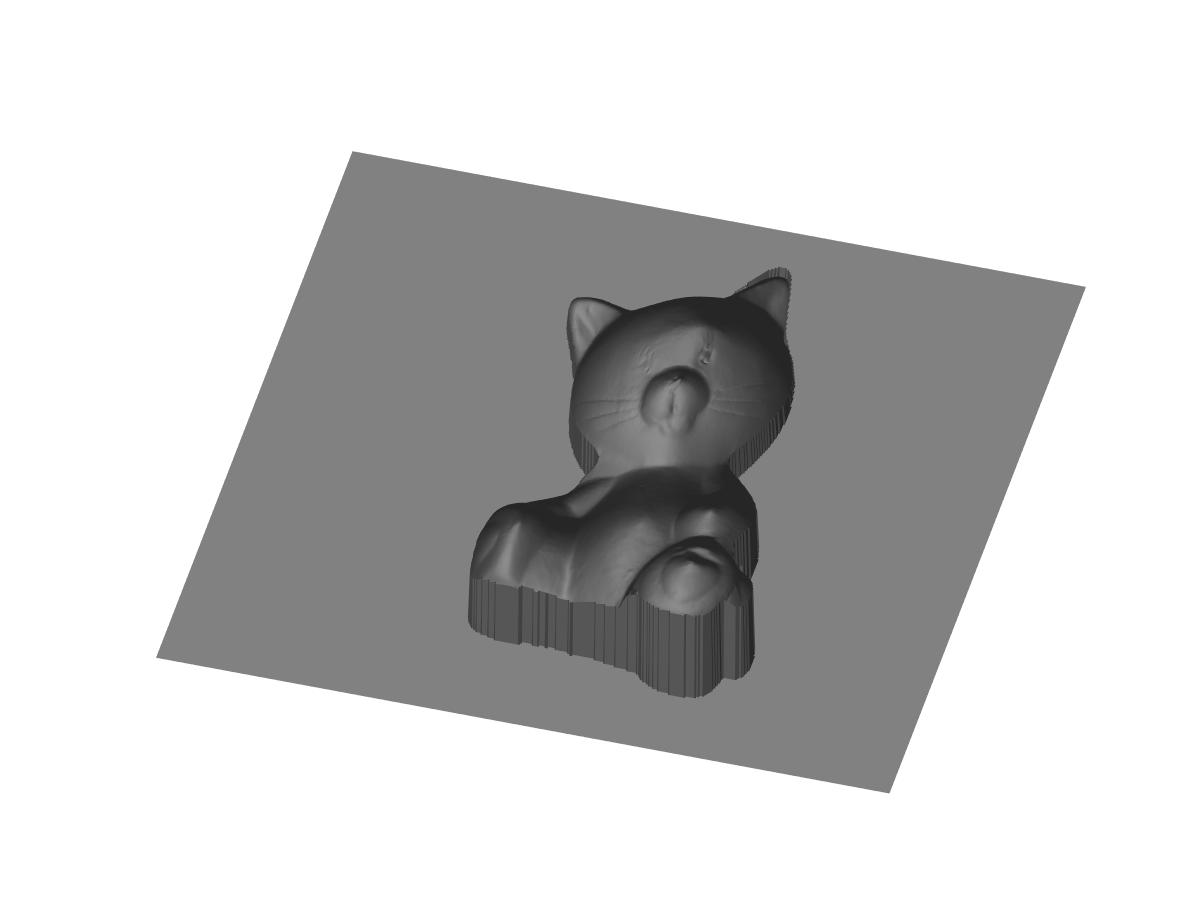}
	\end{subfigure}
	\begin{subfigure}[b]{0.22\textwidth}
		\includegraphics[width=\textwidth]{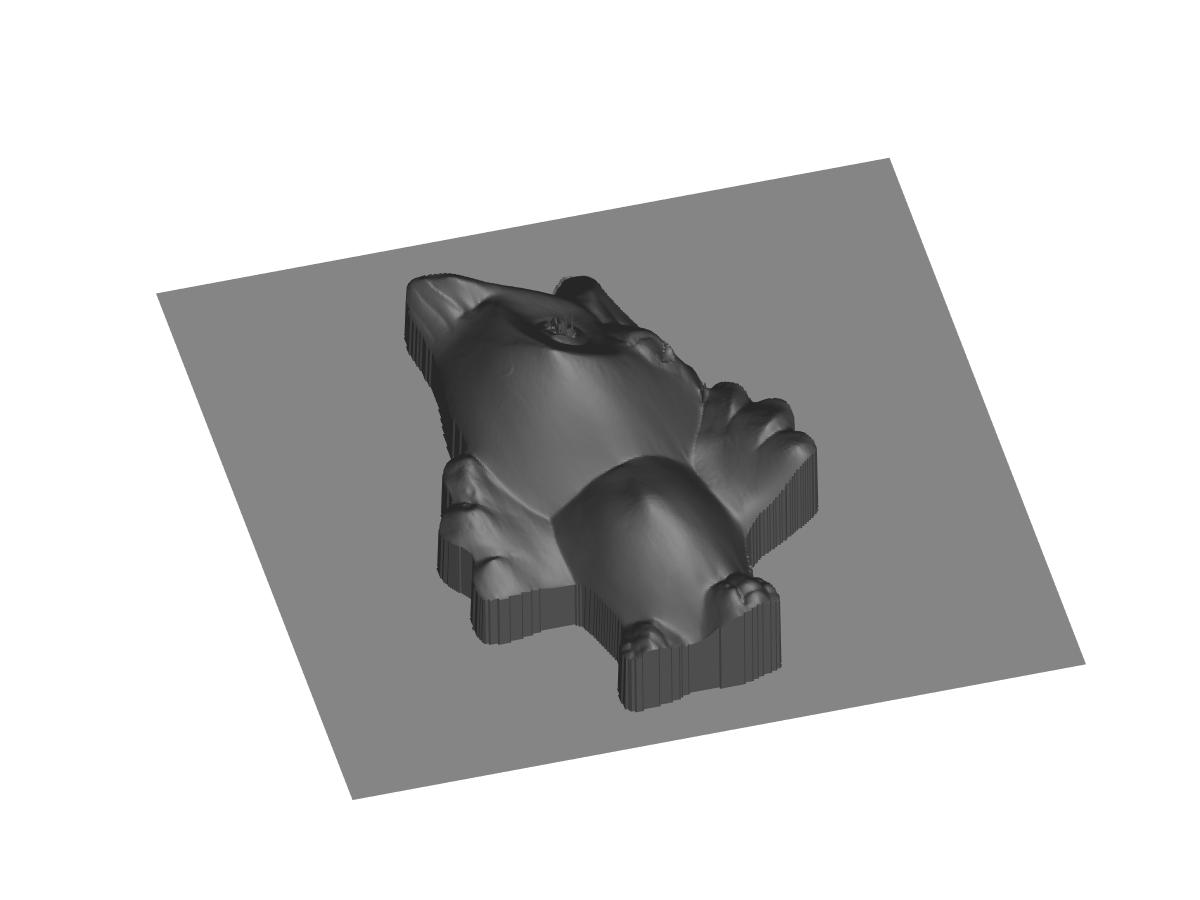}
		\includegraphics[width=\textwidth]{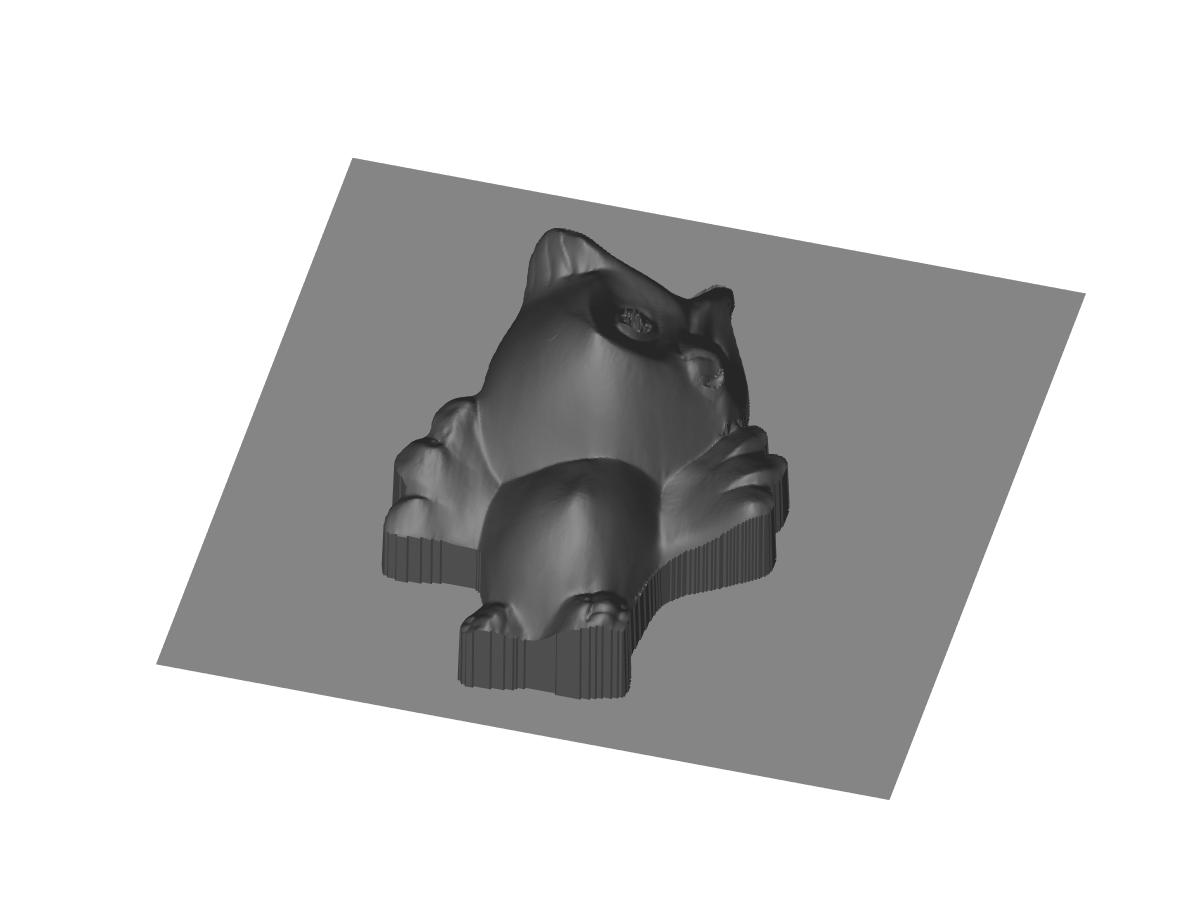}
	\end{subfigure}
	\begin{subfigure}[b]{0.22\textwidth}
		\includegraphics[width=\textwidth]{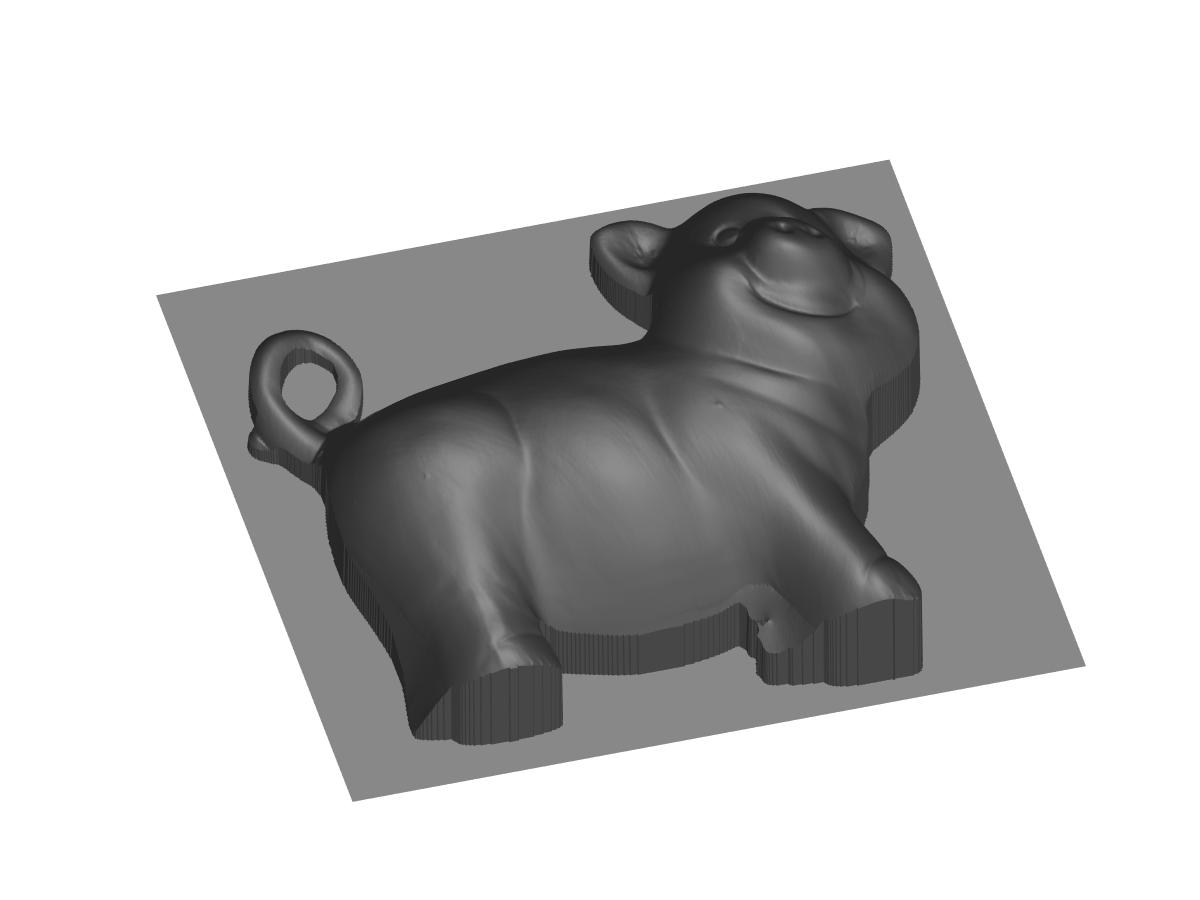}
		\includegraphics[width=\textwidth]{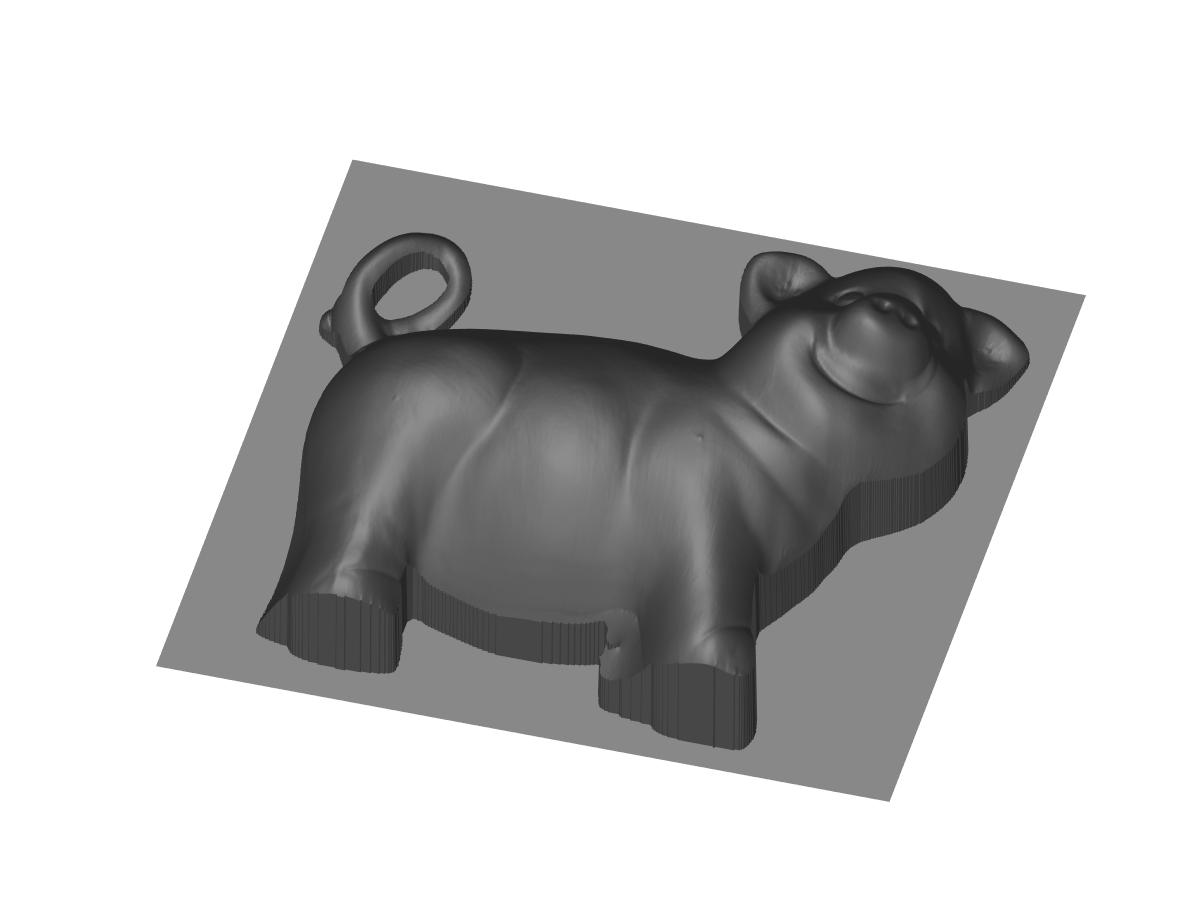}
	\end{subfigure}
	\begin{subfigure}[b]{0.22\textwidth}
		\includegraphics[width=\textwidth]{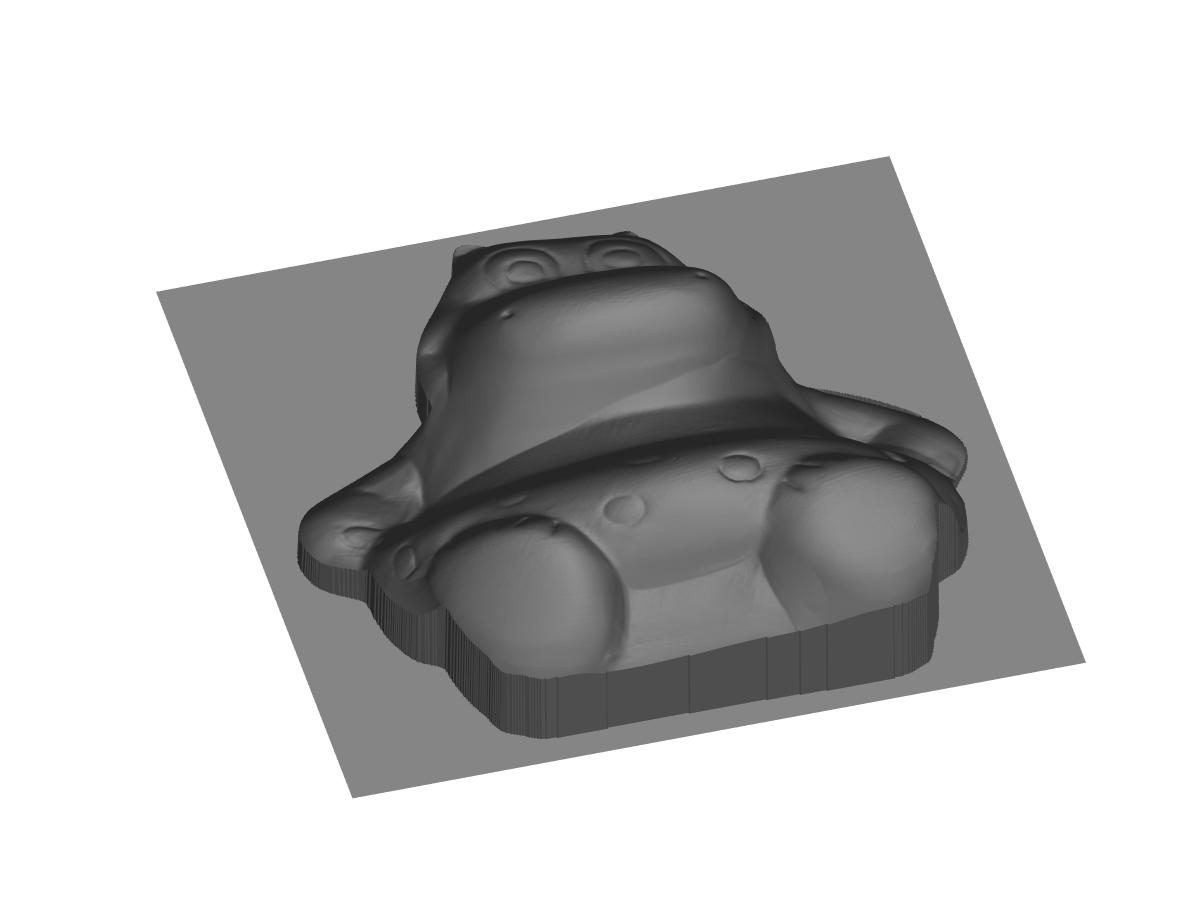}
		\includegraphics[width=\textwidth]{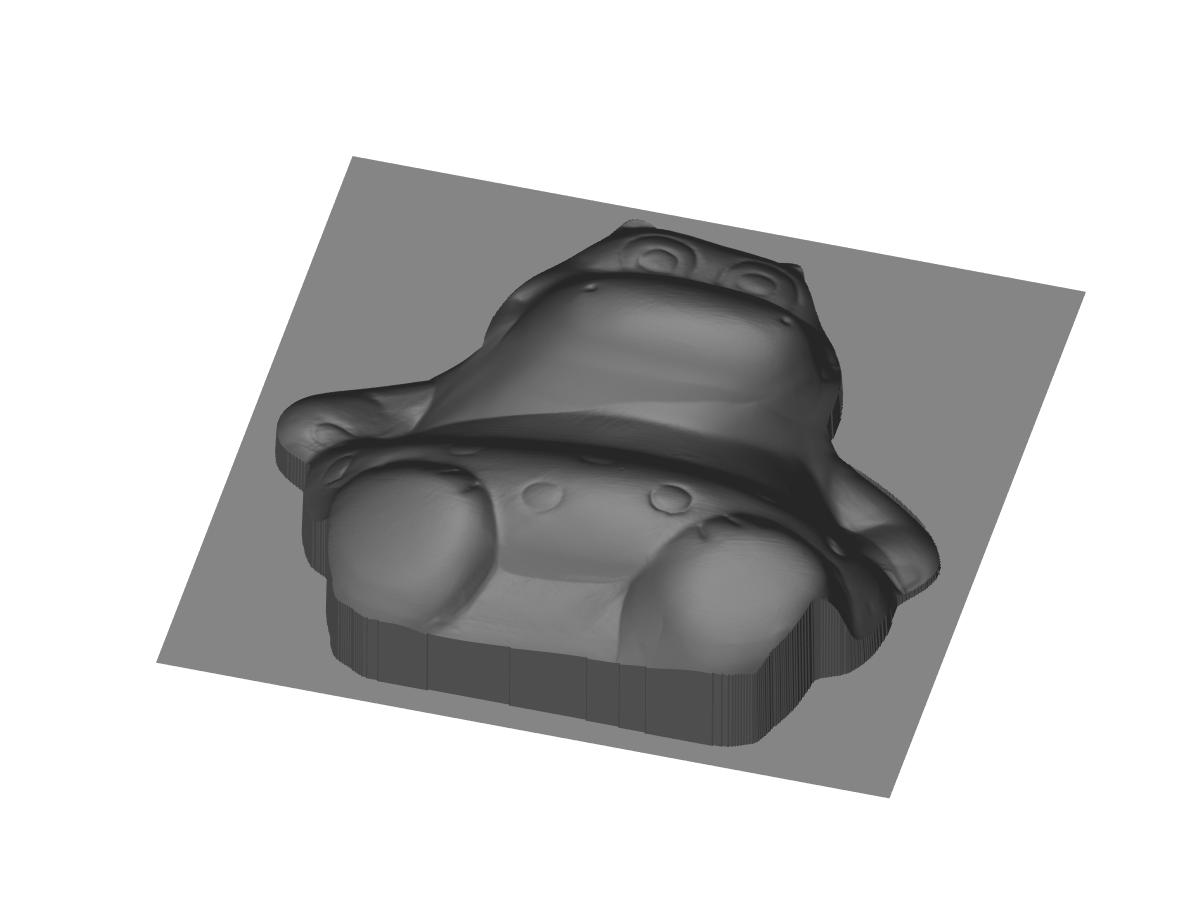}
	\end{subfigure}
	\caption{\small Two views of surfaces reconstructed with Our(MC) algorithm for 4 images. Each column shows two images of surfaces reconstructed on ``Cat", ``Owl", ``Pig" and ``Hippo'' respectively.}
	\label{fig:zview_some}
\end{figure*}

Figure \ref{fig:im} shows the average reconstruction error obtained by Our(MC), RPCA and Baseline on 12 real-world objects over 10 random simulations. We observe that Our(MC) outperforms RPCA on 11 out of 12 objects for 4 images and 10 out of 12 objects for 6 images (and is comparable in 1). With 10 images the average reconstruction error using Our(MC) over all objects and all trials is 4.6\%.  This increases to 8.1\% with four images, and is only 5.4\% with six images. This shows that we have reasonable reconstruction with 4 images and good reconstruction with as few as 6 images.

In Figure \ref{fig:zerr_some} we compare the error in surface reconstruction between Baseline, RPCA, and Our(MC) on some of our real world examples. Figure \ref{fig:zview_some} shows two views of surfaces reconstructed using the Our(MC) algorithm using 4 images, showing reasonable surface reconstruction. These results suggest that our joint approach to enforcing rank and integrability constraints can significantly improve the performance of photometric stereo in the presence of a few images.

In general, we see that incorporating matrix completion into our formulation results in a slight improvement, with Our(MC) somewhat outperforming Our(NC). This indicates that the improvement of our method compared to RPCA or Baseline is mostly due to the joint optimization formulation and not due to matrix completion. We further note that RPCA seems to significantly improve over Baseline.  RPCA is able to identify outliers and use that extra information for better recovery. This also suggests that the robust error function used by RPCA is important. However our integrated approach, which does not have a robust cost function like RPCA, still outperforms RPCA for 4 and 6 images and is almost equal for 8 or 10 images. This shows that an integrated approach is very useful for a small number of images and provides similar gain compared to RPCA for more images. It would be an interesting topic of future work to amend the cost function of Our(MC) to include RPCA's robust handling of error, to see if this further improves its performance.

For an image of size $512 \times 340$ with an object occupying an area of 30K pixels, our algorithm takes 20 minutes on a 2.7 GHz Intel Core i5 machine.

\section{Conclusion and Future Work}

In this paper we have introduced a new low-rank constrained optimization method for solving uncalibrated photometric stereo using fewer images. The key to this approach is to combine rank and integrability constraints in a single optimization problem.  This relies on a novel formulation that exposes both depth and surface normals to the optimization, linking them with an integrability constraint.  We then show how to perform this optimization using a truncated nuclear norm and ADMM. Our joint formulation produces better solutions, compared to other methods that use SVD, for fewer images. We have shown promising results compared to baseline approaches using both real and synthetic examples. We also observe that our method can handle certain degrees of model irregularities as it has outperformed RPCA in synthetic examples with specularities generated using the Phong model.

In the future, it will be interesting to apply the idea of Robust PCA to our formulation. We would also like to extend this work to handle more general lighting configurations, e.g., using spherical harmonic approximations to lighting.

{\small
\bibliographystyle{ieee}
\bibliography{egbib}

\begin{thebibliography}{10}\itemsep=-1pt

\bibitem{ackermann2015survey}
J.~Ackermann and M.~Goesele.
\newblock A survey of photometric stereo techniques.
\newblock {\em Foundations and Trends{\textregistered} in Computer Graphics and
  Vision}, 9(3-4):149--254, 2015.

\bibitem{alldrin}
N.~G. Alldrin, S.~P. Mallick, and D.~J. Kriegman.
\newblock Resolving the generalized bas-relief ambiguity by entropy
  minimization.
\newblock In {\em Computer Vision and Pattern Recognition, 2007. CVPR'07. IEEE
  Conference on}, pages 1--7. IEEE, 2007.

\bibitem{basri}
R.~Basri, D.~Jacobs, and I.~Kemelmacher.
\newblock Photometric stereo with general, unknown lighting.
\newblock {\em International Journal of Computer Vision}, 72(3):239--257, 2007.

\bibitem{gbr}
P.~N. Belhumeur, D.~J. Kriegman, and A.~L. Yuille.
\newblock The bas-relief ambiguity.
\newblock {\em International journal of computer vision}, 35(1):33--44, 1999.

\bibitem{boyd2011distributed}
S.~Boyd, N.~Parikh, E.~Chu, B.~Peleato, and J.~Eckstein.
\newblock Distributed optimization and statistical learning via the alternating
  direction method of multipliers.
\newblock {\em Foundations and Trends{\textregistered} in Machine Learning},
  3(1):1--122, 2011.

\bibitem{cabral}
R.~S. Cabral, F.~Torre, J.~P. Costeira, and A.~Bernardino.
\newblock Matrix completion for multi-label image classification.
\newblock In {\em Advances in Neural Information Processing Systems}, pages
  190--198, 2011.

\bibitem{cai2010shrinkage}
J.-F. Cai, E.~J. Cand\`{e}s, and Z.~Shen.
\newblock A singular value thresholding algorithm for matrix completion.
\newblock {\em SIAM J. on Optimization}, 20(4):1956--1982, 2010.

\bibitem{manu}
M.~Chandraker, S.~Agarwal, and D.~Kriegman.
\newblock Shadowcuts: Photometric stereo with shadows.
\newblock In {\em Computer Vision and Pattern Recognition, 2007. CVPR'07. IEEE
  Conference on}, pages 1--8. IEEE, 2007.

\bibitem{reflec}
M.~K. Chandraker, C.~F. Kahl, and D.~J. Kriegman.
\newblock Reflections on the generalized bas-relief ambiguity.
\newblock In {\em Computer Vision and Pattern Recognition, 2005. CVPR 2005.
  IEEE Computer Society Conference on}, volume~1, pages 788--795. IEEE, 2005.

\bibitem{spec}
O.~Drbohlav and M.~Chaniler.
\newblock Can two specular pixels calibrate photometric stereo?
\newblock In {\em Computer Vision, 2005. ICCV 2005. Tenth IEEE International
  Conference on}, volume~2, pages 1850--1857. IEEE, 2005.

\bibitem{papa12}
P.~Favaro and T.~Papadhimitri.
\newblock A closed-form solution to uncalibrated photometric stereo via diffuse
  maxima.
\newblock In {\em Computer Vision and Pattern Recognition (CVPR), 2012 IEEE
  Conference on}, pages 821--828. IEEE, 2012.

\bibitem{torrence}
A.~S. Georghiades.
\newblock Incorporating the torrance and sparrow model of reflectance in
  uncalibrated photometric stereo.
\newblock In {\em Computer Vision, 2003. Proceedings. Ninth IEEE International
  Conference on}, pages 816--823. Ieee, 2003.

\bibitem{goldstein2014fast}
T.~Goldstein, B.~O'Donoghue, S.~Setzer, and R.~Baraniuk.
\newblock Fast alternating direction optimization methods.
\newblock {\em SIAM Journal on Imaging Sciences}, 7(3):1588--1623, 2014.

\bibitem{Hartley2004}
R.~I. Hartley and A.~Zisserman.
\newblock {\em Multiple View Geometry in Computer Vision}.
\newblock Cambridge University Press, ISBN: 0521540518, second edition, 2004.

\bibitem{hayakawa}
H.~Hayakawa.
\newblock Photometric stereo under a light source with arbitrary motion.
\newblock {\em JOSA A}, 11(11):3079--3089, 1994.

\bibitem{tnn}
Y.~Hu, D.~Zhang, J.~Ye, X.~Li, and X.~He.
\newblock Fast and accurate matrix completion via truncated nuclear norm
  regularization.
\newblock {\em Pattern Analysis and Machine Intelligence, IEEE Transactions
  on}, 35(9):2117--2130, 2013.

\bibitem{data1}
N.~Joshi, I.~Kemelmacher, and I.~Simon.
\newblock Photometric stereo dataset.
\newblock
  url=\url{http://courses.cs.washington.edu/courses/cse455/10wi/projects/project4/},
  2015.

\bibitem{mecca}
R.~Mecca, A.~Tankus, A.~Wetzler, and A.~M. Bruckstein.
\newblock A direct differential approach to photometric stereo with perspective
  viewing.
\newblock {\em SIAM Journal on Imaging Sciences}, 7(2):579--612, 2014.

\bibitem{oh}
T.-H. Oh, H.~Kim, Y.-W. Tai, J.-C. Bazin, and I.~S. Kweon.
\newblock Partial sum minimization of singular values in rpca for low-level
  vision.
\newblock In {\em Computer Vision (ICCV), 2013 IEEE International Conference
  on}, pages 145--152. IEEE, 2013.

\bibitem{attach}
T.~Okabe, I.~Sato, and Y.~Sato.
\newblock Attached shadow coding: Estimating surface normals from shadows under
  unknown reflectance and lighting conditions.
\newblock In {\em Computer Vision, 2009 IEEE 12th International Conference on},
  pages 1693--1700. IEEE, 2009.

\bibitem{wiberg}
T.~Okatani, T.~Yoshida, and K.~Deguchi.
\newblock Efficient algorithm for low-rank matrix factorization with missing
  components and performance comparison of latest algorithms.
\newblock In {\em Computer Vision (ICCV), 2011 IEEE International Conference
  on}, pages 842--849. IEEE, 2011.

\bibitem{papa13}
T.~Papadhimitri and P.~Favaro.
\newblock A new perspective on uncalibrated photometric stereo.
\newblock In {\em Proceedings of the IEEE Conference on Computer Vision and
  Pattern Recognition}, pages 1474--1481, 2013.

\bibitem{phong2}
B.~T. Phong.
\newblock Illumination for computer generated pictures.
\newblock {\em Communications of the ACM}, 18(6):311--317, 1975.

\bibitem{self}
B.~Shi, Y.~Matsushita, Y.~Wei, C.~Xu, and P.~Tan.
\newblock Self-calibrating photometric stereo.
\newblock In {\em Computer Vision and Pattern Recognition (CVPR), 2010 IEEE
  Conference on}, pages 1118--1125. IEEE, 2010.

\bibitem{sunkavalli2010visibility}
K.~Sunkavalli, T.~Zickler, and H.~Pfister.
\newblock Visibility subspaces: Uncalibrated photometric stereo with shadows.
\newblock In {\em European Conference on Computer Vision}, pages 251--264.
  Springer, 2010.

\bibitem{phong1}
P.~Tan.
\newblock Phong reflectance model.
\newblock In {\em Computer Vision}, pages 592--594. Springer, 2014.

\bibitem{iso}
P.~Tan, S.~P. Mallick, L.~Quan, D.~J. Kriegman, and T.~Zickler.
\newblock Isotropy, reciprocity and the generalized bas-relief ambiguity.
\newblock In {\em Computer Vision and Pattern Recognition, 2007. CVPR'07. IEEE
  Conference on}, pages 1--8. IEEE, 2007.

\bibitem{woodham1980photometric}
R.~J. Woodham.
\newblock Photometric method for determining surface orientation from multiple
  images.
\newblock {\em Optical engineering}, 19(1):191139--191139, 1980.

\bibitem{wu}
L.~Wu, A.~Ganesh, B.~Shi, Y.~Matsushita, Y.~Wang, and Y.~Ma.
\newblock Robust photometric stereo via low-rank matrix completion and
  recovery.
\newblock In {\em Computer Vision--ACCV 2010}, pages 703--717. Springer, 2011.

\bibitem{data}
Y.~Xiong, A.~Chakrabarti, R.~Basri, S.~J. Gortler, D.~W. Jacobs, and
  T.~Zickler.
\newblock From shading to local shape.
\newblock {\em Pattern Analysis and Machine Intelligence, IEEE Transactions
  on}, 37(1):67--79, 2015.

\bibitem{yuille}
A.~Yuille and D.~Snow.
\newblock Shape and albedo from multiple images using integrability.
\newblock In {\em Computer Vision and Pattern Recognition, 1997. Proceedings.,
  1997 IEEE Computer Society Conference on}, pages 158--164. IEEE, 1997.

\end{thebibliography}
}

\end{document}